\documentclass[preprint,12pt]{elsarticle}

\usepackage[T1]{fontenc}
\usepackage[utf8]{inputenc}
\usepackage{amssymb}
\usepackage{amsmath}
\usepackage{natbib}
\usepackage{url}
\usepackage{tikz,graphics,color,fullpage,float,epsf,caption,subcaption}
\usepackage{color}
\usepackage{colortbl}
\usepackage{comment}
\usepackage{algorithm}
\usepackage{algpseudocode}
\usepackage{booktabs}
\usepackage{multirow}
\usepackage{array}
\usepackage{placeins}
\usepackage{pdfpages}
\usepackage{enumitem}
\setlist{itemsep=0pt,parsep=0pt,topsep=2pt}

\usetikzlibrary{arrows.meta,positioning,fit,backgrounds,calc,shapes.geometric}

\usepackage{hyperref}
\hypersetup{
    colorlinks=false,        
    linkcolor=blue,         
    filecolor=magenta,      
    urlcolor=cyan,          
    citecolor=blue,         
    bookmarks=true,         
    bookmarksnumbered=true, 
    pdfstartview=FitH       
}

\begin{document}
\begin{frontmatter}



\title{Balancing hierarchical representation and quantization: A performance trade-off in text embeddings\tnoteref{t1}}
\tnotetext[t1]{\copyright{} 2026 The Author(s). This is the authors' version of the open access article published in Information Sciences 763 (2027) 124271, \url{https://doi.org/10.1016/j.ins.2026.124271}, distributed under the CC BY 4.0 license (\url{http://creativecommons.org/licenses/by/4.0/}). Received 3 October 2025; received in revised form 5 October 2026; accepted 6 October 2026; available online 8 October 2026. See the correction notice before the Introduction.}


\author[a]{Kai Ponel\fnmark[1]} 
\affiliation[a]{organization={Ludwig-Maximilian-University Munich},
            addressline={Geschwister-Scholl-Platz 1}, 
            city={Munich},
            country={Germany}}
\author[b,d]{Marco Polignano\corref{cor1}}
\ead{marco.polignano@uniba.it} 
\affiliation[b]{organization={University of Bari Aldo Moro},
            addressline={Via E. Orabona 4}, 
            city={Bari},
            state={Apulia},
            country={Italy}}

\author[c]{Erasmo Purificato\fnmark[1,2]} 
\affiliation[c]{organization={European Commission, Joint Research Centre (JRC)},
    addressline={Via E. Fermi 2749},
    city={Ispra},
    country={Italy}}

\author[ibm]{Aylin Julia Temizkan}
\affiliation[ibm]{organization={IBM Germany},
            addressline={Im Klapperhof 3-5}, 
            city={Cologne},
            country={Germany}}

\author[d]{Marco Levantesi} 
\affiliation[d]{organization={Otto von Guericke University Magdeburg},
            addressline={Universit\"{a}tsplatz 2}, 
            city={Magdeburg},
            country={Germany}}
\author[b]{Giovanni Semeraro} 
\author[d]{Ernesto William De Luca} 

\fntext[1]{The author contributed to this work while affiliated with Otto von Guericke University Magdeburg, Germany.}

\fntext[2]{\textbf{Disclaimer}: The view expressed in this paper is purely that of the author and may not, under any circumstances, be regarded as an official position of the European Commission.}

\cortext[cor1]{Corresponding author at: University of Bari Aldo Moro, Via E. Orabona 4, Bari, Apulia, Italy.}

\begin{abstract}
Hierarchical text embedding models offer variable-length representations, while quantization reduces numerical precision; both aim to improve computational efficiency. This work is an \emph{empirical benchmark study} of how these two orthogonal, already-established techniques interact. We evaluate 38 tasks from the Massive Text Embedding Benchmark (MTEB) using four transformer-based embedding models, covering 1D and 2D hierarchical representations combined with 32-bit, 8-bit, and binary quantization, for 168 configurations per task. We report each efficiency dimension separately: theoretical per-vector storage and end-to-end evaluation time. Under a Friedman test with Nemenyi post-hoc analysis, int8 quantization is statistically indistinguishable from float32 for pair classification and re-ranking and costs at most 1.5 percentage points overall, whereas binary quantization is significantly worse across every task category. Halving the embedding dimension once is not statistically detectable above $m=16$; two successive halvings always are. Combining an 8$\times$ dimensionality reduction with int8 yields a 32$\times$ reduction in per-vector storage at 94.4\% of baseline effectiveness. Because evaluation times were measured once, we treat them as indicative: we find no evidence that quantization speeds up evaluation, whereas sampling from intermediate layers shortens it substantially. Retrieval degrades far faster than every other category, so we report findings per task category.
\end{abstract}

\begin{highlights}
\item A factorial benchmark of Matryoshka truncation and quantization on 38 MTEB tasks.
\item 8-bit embeddings are statistically on par with float32 for several task types.
\item One halving of embedding size is not significant; two successive halvings are.
\item Eightfold truncation plus int8 keeps 94.4\% effectiveness with 32$\times$ less storage.
\item Retrieval is far more sensitive to compression than any other task category.
\end{highlights}

\begin{keyword}
Hierarchical representations \sep Matryoshka representations \sep Embedding quantization \sep Empirical benchmark \sep Text embedding models \sep MTEB
\end{keyword}

\end{frontmatter}

\noindent\fbox{\begin{minipage}{\dimexpr\linewidth-2\fboxsep-2\fboxrule\relax}
\textbf{Correction notice (affiliation of Marco Polignano).} The published version of this article (Information Sciences 763 (2027) 124271, \url{https://doi.org/10.1016/j.ins.2026.124271}) lists the European Commission, Joint Research Centre (JRC) as the second affiliation of Marco Polignano. This is an error in the final manuscript, identified by the authors after publication. The authors asked the publisher to correct it, but the publisher's policy does not allow post-publication changes of this kind. Marco Polignano is not affiliated with the JRC and has no involvement with the JRC. His affiliations are the University of Bari Aldo Moro and Otto von Guericke University Magdeburg, as given in this version. The JRC affiliation applies only to Erasmo Purificato. Apart from this correction and this notice, the content of this version is identical to the published article.
\end{minipage}}

\section{Introduction}
\label{sec:intro}
Learned representations, commonly called embeddings, represent diverse data types (e.g., words, documents) as real-valued vectors, organized such that semantically similar items are positioned close together within the embedding space~\cite{DBLP:journals/jmlr/Bengio12}. Since Word2Vec~\cite{DBLP:journals/corr/abs-1301-3781}, embeddings have advanced further, notably with BERT~\cite{DBLP:conf/naacl/DevlinCLT19}, which marked a shift towards context-dependent embeddings, and with Sentence-BERT~\cite{DBLP:conf/emnlp/ReimersG19} and SimCSE~\cite{DBLP:conf/emnlp/GaoYC21}, designed to learn embeddings for entire sentences. These representations then power a wide spectrum of downstream tasks, including classification, clustering, and retrieval~\cite{DBLP:conf/eacl/MuennighoffTMR23}. Retrieval Augmented Generation (RAG)~\cite{DBLP:conf/nips/LewisPPPKGKLYR020} stands as a prominent example: a typical RAG pipeline involves chunking and embedding documents, retrieving relevant chunks based on similarity to a user query, and then generating a response from a large language model.

The use of embeddings at scale has two primary costs: (1) inference, which involves computing embeddings for new data, and is typically a one-time cost per element, as embeddings can be stored; and (2) downstream tasks, which often involve repeated comparisons between embeddings loaded into memory. An RAG index of one million chunks encoded as 1024-dimensional float32 vectors already occupies about 4~GB. Matryoshka Representations (MRs)~\cite{DBLP:conf/nips/KusupatiBRWSRHC22,DBLP:journals/corr/abs-2402-14776} offer hierarchical embedding storage, enabling the selection of representation size, and with 2D MRs also of the layer, during inference. Additionally, quantization techniques~\cite{JacobKCZTHAK18,shakir2024binary} aim to reduce the bit precision of embedding values, resulting in a smaller memory footprint.

While these techniques have been investigated individually, their combined effect on embedding performance and efficiency remains an open question. This work seeks to address this gap, aiming to answer the following research question: \textit{\textbf{RQ:} ``In the context of transformer-based text embedding models, can hierarchical representations and quantization be combined to improve the computational efficiency of embeddings while maintaining high performance on downstream tasks?''} Our research is guided by four objectives: to evaluate the combined impact of hierarchical representations and quantization across a diverse set of tasks, identifying which task categories are most resilient to dimensionality reduction and precision loss; to quantify the efficiency gains in per-vector storage and end-to-end evaluation time; to determine configurations of dimensionality, numerical precision, and inference layer that balance performance preservation with resource efficiency; and to establish practical guidelines for resource-constrained environments.

\subsection{A worked example}
\label{sec:worked_example}
Consider a retrieval component of a RAG pipeline that stores one million text chunks, each encoded by MBAI-1D (Section~\ref{sec:settings}) into a $d=1024$ float32 vector. Table~\ref{tab:worked_example} traces a single query vector $q$ and two candidate vectors, one relevant ($d_1$), one not ($d_2$), through the two transformations studied in this paper. Truncating to the leading $m=128$ dimensions and then mapping to int8 leaves the \emph{ranking} of $d_1$ above $d_2$ intact while shrinking the index from 4.10~GB to 0.13~GB, a 32$\times$ reduction in stored bytes. Binary quantization compresses the same index to 0.02~GB but maps $q$ and $d_1$ onto \emph{identical} codes, so the distance between them collapses to zero and all ordering information among near neighbors is destroyed. The central empirical question is therefore not whether these transformations save space, but how much task effectiveness each one costs, and on which kinds of tasks.

\begin{table}[!htbp]
\centering
\resizebox{\ifdim\width>\linewidth\linewidth\else\width\fi}{!}{%
\begin{tabular}{@{}llccc@{}}
\toprule
\textbf{Configuration} & \textbf{Vector excerpt (q)} & \textbf{sim$(q,d_1)$ / sim$(q,d_2)$} & \textbf{Bits per vector} & \textbf{Index, 1M vectors} \\
\midrule
float32, $m=1024$ (baseline) & $[0.62, -0.41, 0.33, 0.18]$ & $+1.00$ / $-0.78$ & 32{,}768 & 4.10 GB \\
float32, $m=128$ (1D MR)     & $[0.62, -0.41, 0.33, 0.18]$ & $+1.00$ / $-0.77$ & 4096  & 0.51 GB \\
int8, $m=128$ (MR + int8)    & $[121, -81, 64, 35]$        & $+1.00$ / $-0.78$ & 1024  & 0.13 GB \\
binary, $m=128$ (MR + 1-bit) & $[1, 0, 1, 1]$              & Hamming 0 / 3     & 128      & 0.02 GB \\
\bottomrule
\end{tabular}}
\caption{Worked example of the two transformations studied in this paper, applied to a query and two candidate chunks. Similarity is cosine similarity for the real-valued and integer rows and Hamming distance for the binary row. Vector values are an illustrative four-dimensional excerpt; bit counts are exact for $d=1024$ and $m=128$.}
\label{tab:worked_example}
\end{table}

\subsection{Contributions and positioning}
\label{sec:contributions}
We stress that this paper is an \textbf{empirical benchmark and guideline study}, not a proposal for a new model, training objective, or compression algorithm. Specifically:
\begin{enumerate}
    \item \textbf{A controlled factorial benchmark.} We evaluate the full cross-product of hierarchical size, numerical precision, and (for 2D models) inference layer, with 168 configurations per task, over 38 MTEB tasks and four transformer-based embedding models.
    \item \textbf{Disaggregated efficiency accounting.} We separate theoretical per-vector storage from measured end-to-end evaluation wall-clock time, and we show that the two do \emph{not} move together.
    \item \textbf{Statistical validation.} We subject the results to Friedman, Nemenyi, and Wilcoxon signed-rank testing, which shows, for example, that a single halving of the embedding dimension is not statistically detectable at any size above $m=16$, whereas two successive halvings always are.
    \item \textbf{Task-category-resolved guidelines.} Because different MTEB categories react very differently to compression, we report and recommend per category, and we identify retrieval as the category that constrains aggressive optimization.
\end{enumerate}

\paragraph{Structure of the paper} Section~\ref{sec:background} reviews hierarchical text representations and quantization strategies. Section~\ref{sec:methodology} presents the methodology and Section~\ref{sec:settings} details the experimental settings. Section~\ref{sec:results} reports the experimental results. Section~\ref{sec:discussion} derives practical deployment guidelines. Section~\ref{sec:limitations} states the limitations of the study and Section~\ref{sec:conclusion} concludes.

\section{Background and related work}
\label{sec:background}
This section reviews the two lines of work our study combines, hierarchical text representations and embedding quantization. What has not previously been established is how they behave \emph{in combination}: work on hierarchical representations varies dimensionality at full precision, while work on quantization varies precision at full dimensionality. Practitioners nonetheless apply both simultaneously, on the untested assumption that two orthogonal compression mechanisms compose without interacting. Whether that assumption holds is an empirical question that requires a factorial design to answer.

\subsection{Hierarchical text representation}
Traditional embeddings do not allow for flexibility when it comes to the dimensionality of the representation: all the $d$ dimensions must be used in downstream tasks to avoid significant drops in performance~\cite{DBLP:conf/nips/YinS18}, which has led to various attempts to develop more flexible representations~\cite{DBLP:journals/csur/LiGTTHLZ24}. Principal Component Analysis (PCA)~\cite{DBLP:conf/rep4nlp/GuptaGRB19,DBLP:conf/rep4nlp/RaunakGM19} projects high-dimensional embeddings onto a lower-dimensional subspace, aiming to preserve as much variance as possible. It is fast and straightforward, but it is inherently a linear transformation, which limits its ability to capture complex semantic structures, and the projection must be fitted and stored for each corpus. Jointly-optimized subnetworks~\cite{DBLP:conf/iclr/CaiGWZH20,DBLP:journals/corr/abs-1812-08928} involve training multiple interconnected subnetworks within a larger framework, enabling the system to dynamically adjust the dimensionality of the embeddings. While these methods offer greater flexibility, training such subnetworks requires substantial computational resources.

\textbf{Matryoshka Representations.} Matryoshka Representations (MRs), inspired by Russian nesting dolls, were introduced by Kusupati et al.~\cite{DBLP:conf/nips/KusupatiBRWSRHC22}: for every $m \in \mathbb{M}$, the first $m$ dimensions of the embedding vector $z_{1:m} \in \mathbb{R}^{m}$ can be used as a general purpose representation of the input data. $\mathbb{M}$ is composed of consistent reductions by half from the full embedding size; for $d = 2048$, for example, $\mathbb{M} = \{2048, 1024, \dots, 8\}$. MRs simultaneously optimize multiple loss functions across the specified dimensions, which encourages the model to prioritize essential information in the leading dimensions. This method requires minimal modifications to existing training pipelines, imposes no additional cost during inference, and matches or outperforms independently trained low-dimensional representations~\cite{DBLP:conf/nips/KusupatiBRWSRHC22}.

\textbf{2D Matryoshka Representations.} Although MRs greatly improve the efficiency of learned representations, the forward pass required to compute these embeddings is still expensive, since the model has to process the input through all layers. 2D MRs~\cite{DBLP:journals/corr/abs-2402-14776} allow the user to choose both layers and the size of the output embeddings at inference time. The authors of 2D MRs based their work on a pre-trained BERT model~\cite{DBLP:conf/naacl/DevlinCLT19} with [CLS] pooling~\cite{DBLP:conf/emnlp/GaoYC21,DBLP:conf/acl/LiL24}:
\begin{equation}
X_{1:n}^{m} = \text{BERT}_{\text{cls}}^{1:n}(x)_{1:m} \in \mathbb{R}^{m}
\end{equation}
where $n \in [1, N]$ denotes the $n$-th layer of the $N$-layer Transformer backbone and $m$ is the dimension of the embedding vector. To address the low quality of the embeddings obtained from shallow layers, the authors align them with the final layer through divergence minimization.

\subsection{Quantization strategies}
In deep learning, quantization applies to two distinct areas. Model parameter quantization focuses on reducing the precision of weights and activations, through post-training quantization or quantization-aware training, to decrease memory usage and improve computational speed~\cite{DBLP:journals/sigmobile/LinTTYXH24,DBLP:conf/aaai/YaoW0YH24}. In contrast, embedding quantization targets the floating-point vectors that encode semantic relationships between words or sentences, whose memory overhead dominates in large-scale applications involving billions of vectors.

The simplest approach involves floating-point downsampling to Float16 or BFloat16, which achieves a straightforward 2$\times$ memory reduction, often with negligible accuracy loss~\cite{abs250500105}. Converting embeddings to int8 reduces memory usage by 4$\times$, and integer arithmetic is both faster and more power-efficient than floating-point operations~\cite{JacobKCZTHAK18}. The most aggressive approach involves binary (1-bit) quantization, which yields a 32$\times$ memory reduction relative to float32; in practice, search over binary codes with Hamming distance can achieve approximately a 25$\times$ speedup compared to float32 search on the CPU, but binary quantization inevitably loses semantic nuance~\cite{shakir2024binary}. Alternative approaches include vector quantization methods, where embeddings are represented as discrete codes drawn from a learned codebook. Product quantization (PQ)~\cite{DBLP:journals/pami/JegouDS11} exemplifies this paradigm by dividing the embedding space into lower-dimensional subspaces and quantizing each subspace independently; together with its optimized variant (OPQ)~\cite{DBLP:conf/cvpr/GeHK013}, it is the standard compression method in approximate nearest neighbor search libraries such as Faiss~\cite{douze2024faiss}.

Since quantization and Matryoshka are orthogonal techniques, we can first shrink an embedding via Matryoshka, then apply integer or binary quantization, compounding the memory savings; Huerga-P\'erez et al.~\cite{abs250500105} study such combinations for RAG storage. Table~\ref{tab:comp} summarizes the trade-offs of the main methods. The accuracy figures reported for them originate from different papers, corpora, and evaluation protocols, and therefore support qualitative positioning only; a controlled comparison of PQ, PCA, and float16 is outside the scope of this work (Section~\ref{sec:limitations}). What Table~\ref{tab:comp} does show is a structural distinction: PCA and PQ obtain their compression from a transformation that must first be \emph{fitted} to a corpus and then stored and re-fitted under distributional drift, whereas hierarchical truncation is a property of how the model was trained and costs nothing at inference, so one stored vector serves every supported budget.

\begin{table}[!htbp]
\centering
\small
\resizebox{\ifdim\width>\linewidth\linewidth\else\width\fi}{!}{%
\begin{tabular}{@{}>{\raggedright\arraybackslash}p{2.9cm}cc >{\raggedright\arraybackslash}p{4.5cm} >{\raggedright\arraybackslash}p{4.5cm}@{}}
\toprule
\textbf{Method} & \textbf{Compression} & \textbf{Fitting} & \textbf{Pros} & \textbf{Cons} \\
 & \textbf{vs float32} & \textbf{required} & & \\
\midrule
Float16 / BFloat16~\cite{abs250500105} & 2$\times$ & No & Plug-and-play; hardware-native on modern GPUs; small reported accuracy loss & Modest compression; bfloat16 trades precision for range \\
\addlinespace
Int8 scalar quantization~\cite{JacobKCZTHAK18} & 4$\times$ & Value ranges & Integer arithmetic is fast and energy-efficient; widely supported & Ranges must be estimated from data; outlier values reduce the resolution \\
\addlinespace
Binary quantization~\cite{shakir2024binary} & 32$\times$ & No & Extreme compression; fast Hamming-distance search on CPU; data-independent & Large semantic loss; close neighbors collapse to identical codes; usually needs rescoring \\
\addlinespace
Product quantization (PQ, OPQ)~\cite{DBLP:journals/pami/JegouDS11,DBLP:conf/cvpr/GeHK013} & 8--32$\times$ & Codebook & Strong compression-accuracy trade-off; mature nearest-neighbor search support & Codebook must be trained on in-domain data and retrained under drift; more complex distance computation \\
\addlinespace
PCA projection~\cite{DBLP:conf/rep4nlp/GuptaGRB19,DBLP:conf/rep4nlp/RaunakGM19} & 2--10$\times$ & Projection & Cheap to fit and apply; orthogonal to precision reduction & Linear, so it misses non-linear structure; projection must be stored and refitted per corpus \\
\midrule
1D/2D MR $+$ int8 (this study) & 32$\times$ at $m = d/8$ & No & Truncation is free at inference; one stored vector serves every size; composes with precision reduction & Requires a model trained with an MR objective; storage gains bring little speed gain; retrieval degrades at small sizes \\
\bottomrule
\end{tabular}}
\caption{Pros and cons of the main embedding compression techniques. Compression ratios are relative to float32 storage of the same number of components; the last row summarizes the configuration measured in this study.}
\label{tab:comp}
\end{table}

\section{Methodology}
\label{sec:methodology}
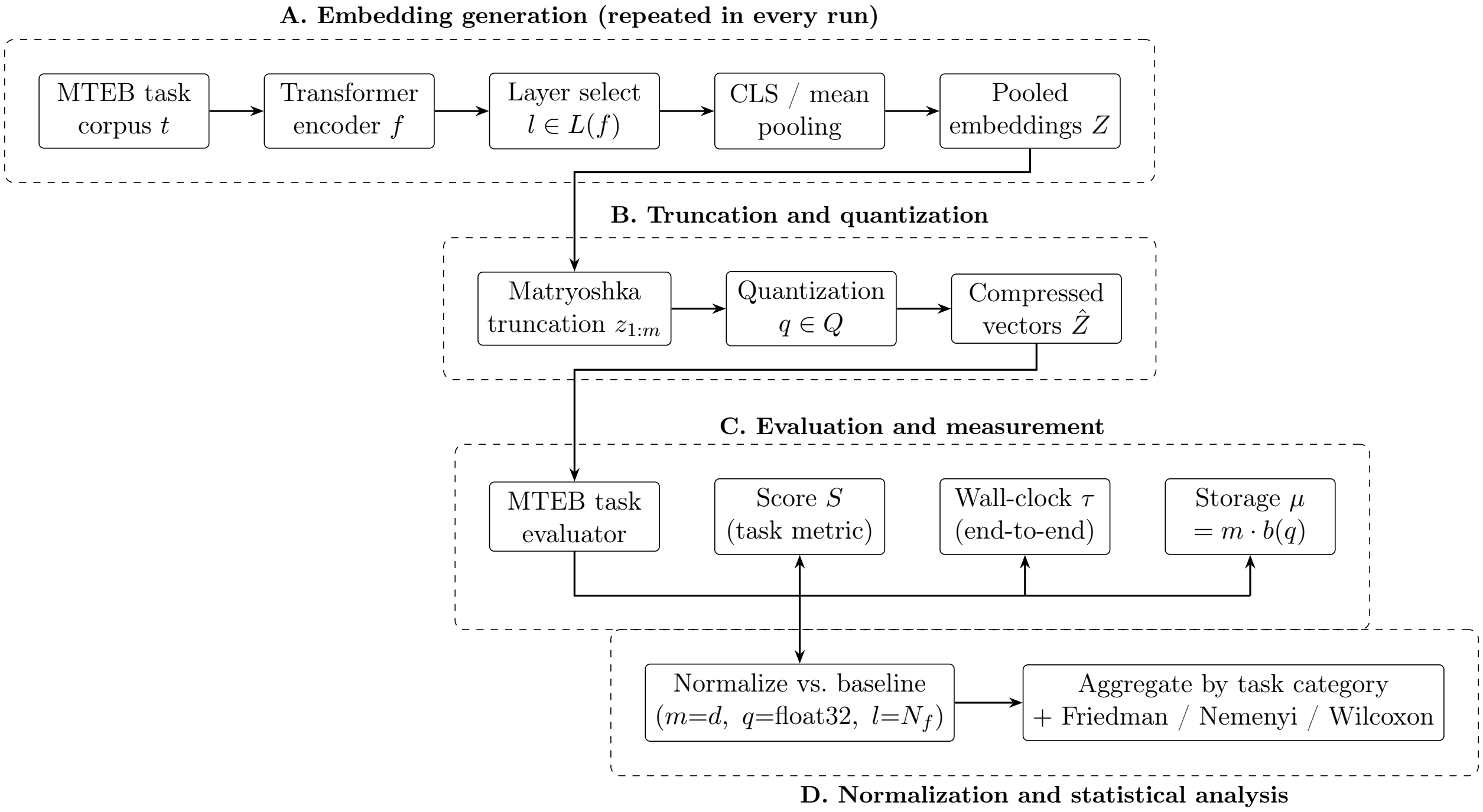
\begin{figure}[!htbp]
\centering
\resizebox{\ifdim\width>\linewidth\linewidth\else\width\fi}{!}{%
\begin{tikzpicture}[
  font=\small,
  box/.style={draw,rounded corners=2pt,minimum height=9mm,minimum width=25mm,align=center,fill=white},
  stage/.style={draw,dashed,rounded corners=4pt,inner sep=5mm},
  ar/.style={-{Stealth[length=2mm]},thick},
  lbl/.style={font=\scriptsize\itshape}
]
\node[box] (txt)  {MTEB task\\corpus $t$};
\node[box,right=8mm of txt]  (enc) {Transformer\\encoder $f$};
\node[box,right=8mm of enc]  (lay) {Layer select\\$l \in L(f)$};
\node[box,right=8mm of lay]  (pool){CLS / mean\\pooling};
\node[box,right=8mm of pool] (cache){Pooled\\embeddings $Z$};
\draw[ar] (txt)--(enc); \draw[ar] (enc)--(lay); \draw[ar] (lay)--(pool); \draw[ar] (pool)--(cache);

\node[box,below=18mm of lay]   (trunc) {Matryoshka\\truncation $z_{1:m}$};
\node[box,right=8mm of trunc]  (quant) {Quantization\\$q \in Q$};
\node[box,right=8mm of quant]  (store) {Compressed\\vectors $\hat{Z}$};
\draw[ar] (cache) -- ++(0,-9mm) -| (trunc);
\draw[ar] (trunc)--(quant); \draw[ar] (quant)--(store);

\node[box,below=20mm of trunc] (eval) {MTEB task\\evaluator};
\node[box,right=8mm of eval]   (perf) {Score $S$\\(task metric)};
\node[box,right=8mm of perf]   (time) {Wall-clock $\tau$\\(end-to-end)};
\node[box,right=8mm of time]   (mem)  {Storage $\mu$\\$= m \cdot b(q)$};
\draw[ar] (store) -- ++(0,-9mm) -| (eval);
\coordinate (b1) at ($(perf.south)+(0,-6mm)$);
\coordinate (b3) at ($(mem.south)+(0,-6mm)$);
\draw[thick] (eval.south) |- (b1);
\draw[thick] (b1) -- (b3);
\draw[ar] (b1) -- (perf.south);
\draw[ar] ($(time.south)+(0,-6mm)$) -- (time.south);
\draw[ar] (b3) -- (mem.south);

\node[box,below=16mm of perf,minimum width=36mm] (norm) {Normalize vs.\ baseline\\$(m{=}d,\ q{=}\textrm{float32},\ l{=}N_f)$};
\node[box,right=10mm of norm,minimum width=42mm] (stat) {Aggregate by task category\\$+$ Friedman / Nemenyi / Wilcoxon};
\draw[ar] ($(perf.south)+(0,-6mm)$) -- (norm.north);
\draw[ar] (norm)--(stat);

\begin{scope}[on background layer]
  \node[stage,fit=(txt)(cache),label={[font=\footnotesize\bfseries]above:A. Embedding generation (repeated in every run)}] {};
  \node[stage,fit=(trunc)(store),label={[font=\footnotesize\bfseries]above:B. Truncation and quantization}] {};
  \node[stage,fit=(eval)(mem)(b1)(b3),label={[font=\footnotesize\bfseries]above:C. Evaluation and measurement}] {};
  \node[stage,fit=(norm)(stat),label={[font=\footnotesize\bfseries]below:D. Normalization and statistical analysis}] {};
\end{scope}
\end{tikzpicture}}
\caption{Component flow of the evaluation pipeline. Stages A to C are executed anew for every configuration $(f,l,m,q)$ within one MTEB run; embeddings are not cached across configurations, so the wall-clock time $\tau$ recorded for each task covers embedding generation, truncation and quantization, and task evaluation. Stage D normalizes every measurement against the model's own baseline configuration and feeds the aggregated results into the statistical analysis in Section~\ref{sec:stats}.}
\label{fig:architecture}
\end{figure}

We use datasets from the Massive Text Embedding Benchmark (MTEB)~\cite{DBLP:conf/eacl/MuennighoffTMR23}, which are considered standard in the literature, and we test each combination of hierarchical size and quantization technique for each model, together with the inference layer for the 2D model. The overall process is given step by step in Algorithm~\ref{alg:main}, and the corresponding component flow is shown in Fig.~\ref{fig:architecture}.

\subsection{Notation}
\label{sec:notation}
Table~\ref{tab:notation} collects the symbols used throughout the paper. Three factors are varied independently: the hierarchical size $m$, the quantization technique $q$, and, for 2D models only, the inference layer $l$. A \emph{configuration} is a triple $(m,q,l)$ evaluated for a given model on a given task.

\begin{table}[!htbp]
\centering
\small
\resizebox{\ifdim\width>\linewidth\linewidth\else\width\fi}{!}{%
\begin{tabular}{@{}ll l@{}}
\toprule
\textbf{Symbol} & \textbf{Domain} & \textbf{Meaning} \\
\midrule
$F$, $f$ & n/a &  Set of evaluated embedding models; an individual model \\
$N_f$ & $\mathbb{N}$ & Number of transformer layers of model $f$ \\
$d$ & $\mathbb{N}$ & Default (maximum) output dimensionality of a model \\
$m$ & $\mathbb{M}$ & Hierarchical (truncated) embedding size, $m \le d$ \\
$\mathbb{M}$ & $\subset \mathbb{N}$ & Set of hierarchical sizes, successive halvings of $d$ \\
$z$ & $\mathbb{R}^{d}$ & Full embedding vector produced by a model \\
$z_{1:m}$ & $\mathbb{R}^{m}$ & Matryoshka prefix: leading $m$ components of $z$ \\
$q$ & $Q$ & Quantization technique \\
$Q$ & n/a &  $\{\textrm{float32},\ \textrm{int8},\ \textrm{binary}\}$ \\
$b(q)$ & $\{32,8,1\}$ & Bits per stored component under technique $q$ \\
$l$ & $L(f)$ & Transformer layer used for inference \\
$L(f)$ & n/a &  $\{N_f\}$ for 1D models; $\{24,20,16,12\}$ for the 2D model \\
$T$, $t$ & n/a &  Set of MTEB tasks; an individual task \\
$c$ & $C$ & Task category (clustering, retrieval, STS, \dots), $|C| = 7$ \\
$S(f,t,m,q,l)$ & $\mathbb{R}$ & Absolute task score under the task's own main metric \\
$\hat{S}$ & $\mathbb{R}^{+}$ & Relative performance, $S / S_{\text{base}}$ (Eq.~\ref{eq:relperf}) \\
$\tau(f,t,m,q,l)$ & $\mathbb{R}^{+}$ & Absolute end-to-end task evaluation time (seconds) \\
$\hat{\tau}$ & $\mathbb{R}^{+}$ & Relative evaluation time, $\tau / \tau_{\text{base}}$ \\
$\mu$ & $\mathbb{N}$ & Theoretical per-vector storage in bits, $\mu = m \cdot b(q)$ \\
$\text{base}$ & n/a &  Reference configuration $(m=d,\ q=\textrm{float32},\ l=N_f)$ \\
\midrule
$k$ & $\mathbb{N}$ & Number of levels of the factor under test (Friedman) \\
$N_b$ & $\mathbb{N}$ & Number of matched blocks (Friedman) \\
$\bar{R}_j$ & $\mathbb{R}$ & Mean rank of level $j$; rank 1 denotes the best level \\
$\mathrm{CD}$ & $\mathbb{R}^{+}$ & Nemenyi critical difference at $\alpha = 0.05$ \\
$W$ & $[0,1]$ & Kendall's coefficient of concordance \\
\bottomrule
\end{tabular}}
\caption{Notation used throughout the paper.}
\label{tab:notation}
\end{table}

\subsection{Evaluation pipeline architecture}
\label{sec:architecture}
Every configuration $(f,l,m,q)$ is evaluated in a separate MTEB run, in which the benchmark calls the embedding model on the texts of each task, so embeddings are never cached across configurations. Each run repeats the forward pass through the first $l$ transformer layers, pools the output, keeps its leading $m$ components and quantizes them before the task evaluator computes the score, so the recorded evaluation time always includes embedding generation. Because truncation and quantization are applied after pooling, the encoder computation is identical for all hierarchical sizes and precisions at a given layer. For traditional 1D MRs, there is only one inference layer, which is the final layer of the model, whereas for 2D MRs there are multiple inference layers that can serve as the output layer of the respective submodel, so the number of runs per task is
\begin{equation}
|\mathbb{M}| \times |Q| \times \left(|F_{1D}| + |F_{2D}| \times |L|\right),
\end{equation}
where $|F_{1D}|$ is the number of models supporting 1D hierarchical representations, $|F_{2D}|$ is the number of models supporting 2D MRs, and $|L|$ is the number of inference layers for 2D MRs.

\begin{algorithm}[!htbp]
\footnotesize
\caption{Factorial evaluation of hierarchical representation $\times$ quantization}
\label{alg:main}
\begin{algorithmic}[1]
\Require Models $F$; tasks $T$; quantization set $Q=\{\textrm{float32},\textrm{int8},\textrm{binary}\}$;
         hierarchical sizes $\mathbb{M}(f)$; inference layers $L(f)$
\Ensure Result set $\mathcal{R}$ of tuples $(f,t,l,m,q,\hat{S},\hat{\tau},\mu)$
\State $\mathcal{R} \gets \emptyset$
\For{each model $f \in F$}
  \State $L(f) \gets \{N_f\}$ \textbf{if} $f$ is a 1D MR model \textbf{else} $\{24,20,16,12\}$
  \For{each configuration $(q,l,m) \in Q \times L(f) \times \mathbb{M}(f)$}
    \State $g \gets \textsc{Configure}(f,l,m,q)$ \Comment{first $l$ layers; output cut to $m$, quantized to $q$}
    \For{each task $t \in T$} \Comment{one MTEB run per configuration}
      \State $\tau_0 \gets \textsc{Clock}()$
      \State $S \gets \textsc{EvaluateMTEB}(t,\,g)$ \Comment{encodes the texts of $t$ with \textsc{Embed}}
      \State $\tau \gets \textsc{Clock}() - \tau_0$ \Comment{includes embedding generation}
      \State $\mu \gets m \cdot b(q)$ \Comment{theoretical per-vector storage, bits}
      \State $\mathcal{R} \gets \mathcal{R} \cup \{(f,t,l,m,q,S,\tau,\mu)\}$
    \EndFor
  \EndFor
\EndFor
\Statex
\State \textbf{Normalization:}
\For{each $(f,t)$ pair in $\mathcal{R}$}
  \State $(S_{\text{base}},\tau_{\text{base}}) \gets$ entry with $m{=}d,\ q{=}\textrm{float32},\ l{=}N_f$
  \State $\hat{S} \gets S / S_{\text{base}}$ \textbf{;} \quad $\hat{\tau} \gets \tau / \tau_{\text{base}}$
         \Comment{Eq.~\ref{eq:relperf}}
\EndFor
\Statex
\State \textbf{Aggregation and analysis:}
\State Group $\hat{S},\hat{\tau}$ by task category $c \in C$ and report per category
\State \textsc{Friedman}$(\hat{S})$ over each factor in $\{q,\ m,\ l\}$, blocking on the other two and on $c$
\State \textbf{if} the null is rejected \textbf{then} \textsc{Nemenyi} post-hoc $+$ \textsc{Wilcoxon} signed-rank with Holm correction
\Statex
\Function{Embed}{$g$, texts} \Comment{repeated in every run, never cached}
  \State $Z \gets$ pooled output of the first $l$ layers of $f$
  \State $Z_m \gets Z[:,\,1{:}m]$ \Comment{Matryoshka truncation}
  \State \Return $\textsc{Quantize}(Z_m,\,q)$
\EndFunction
\Function{Quantize}{$Z_m$, $q$} \Comment{Sentence-Transformers \texttt{quantize\_embeddings}}
  \If{$q = \textrm{float32}$} \State \Return $Z_m$
  \ElsIf{$q = \textrm{int8}$}
    \State for each dimension $j$: $lo_j \gets \min_i Z_{ij}$, $hi_j \gets \max_i Z_{ij}$, $s_j \gets (hi_j - lo_j)/255$
    \State \Return $\operatorname{trunc}\!\left((Z_m - lo)/s - 128\right)$ as signed 8-bit integers
  \ElsIf{$q = \textrm{binary}$}
    \State \Return $\mathbf{1}[Z_m > 0]$ packed eight to a byte \Comment{evaluated as numeric vectors}
  \EndIf
\EndFunction
\end{algorithmic}
\end{algorithm}

\subsection{Evaluation metrics}
\label{sec:metrics}
For each configuration we collect three quantities, which we keep separate throughout the paper.

\paragraph{1. Relative performance} The task's own main metric (Section~\ref{sec:benchmark}), normalized by the score of the same model's baseline configuration on the same task:
\begin{equation}
\hat{S}(f,t,m,q,l) \;=\; \frac{S(f,t,m,q,l)}{S(f,t,\,d,\,\textrm{float32},\,N_f)}
\label{eq:relperf}
\end{equation}
The baseline configuration uses the highest-dimensional embeddings, no quantization, and the final layer for inference. Normalization is applied \emph{per task}, before any aggregation, so that tasks measured on different scales contribute comparably.

\paragraph{2. Evaluation time} We measure the \emph{end-to-end wall-clock time to complete one MTEB task}, denoted $\tau$, and normalize it as $\hat{\tau} = \tau / \tau_{\text{base}}$. It comprises embedding generation, truncation, and quantization; the task-specific downstream computation (the similarity search for retrieval and re-ranking, the logistic-regression probe for classification, the $k$-means fit for clustering, and the pairwise similarity computation for the other categories); and the MTEB harness overhead. The encoder forward pass dominates the evaluation time of most task categories and is identical for every hierarchical size and precision at a given layer, so differences between such configurations are diluted by the common encoding cost, whereas changing the inference layer changes the encoder cost itself. Our timing results are therefore \emph{benchmark throughput} figures rather than per-query inference latency.

\paragraph{3. Theoretical per-vector storage} The number of bits required to store one embedding vector:
\begin{equation}
\mu \;=\; m \times b(q), \qquad b(\textrm{float32}) = 32,\ \ b(\textrm{int8}) = 8,\ \ b(\textrm{binary}) = 1
\label{eq:memory}
\end{equation}
This quantity is \emph{exact and analytic}, not measured. It does not transfer one-to-one to resident system memory, which also includes the model weights, the framework overhead, any index structure, and, for two-stage pipelines, the higher-precision copies retained for rescoring; in a system dominated by vector storage, such as a large RAG index, $\mu$ is a good proxy for the dominant term. Where we report memory reductions in this paper, we mean reductions in $\mu$.

\subsection{Benchmark and tasks}
\label{sec:benchmark}
We use MTEB~\cite{DBLP:conf/eacl/MuennighoffTMR23}, whose public leaderboard\footnote{\url{https://huggingface.co/spaces/mteb/leaderboard}} ranks models across task categories. MTEB includes 58 benchmarks across 8 different task categories; we use the seven categories below and exclude bitext mining, which is multilingual.
\begin{itemize}
    \item \textbf{Classification:} assigning a text to a predefined category after training a logistic regression model; \textbf{accuracy}.
    \item \textbf{Clustering:} grouping texts with the k-means algorithm; \textbf{V-Measure}~\cite{DBLP:conf/emnlp/RosenbergH07}.
    \item \textbf{Pair Classification:} assigning a label to a pair of texts, indicating their relationship, based on their cosine similarity; \textbf{Average Precision (AP)}~\cite{DBLP:conf/sigir/AslamY06}.
    \item \textbf{Re-ranking:} given a query and a set of relevant and irrelevant texts, ranking the texts so that the most relevant ones appear higher; \textbf{Mean Average Precision (MAP)}.
    \item \textbf{Retrieval:} retrieving the most relevant documents from a corpus in response to a query; \textbf{Normalized Discounted Cumulative Gain (NDCG)}~\cite{DBLP:journals/tois/JarvelinK02}.
    \item \textbf{Semantic Textual Similarity (STS):} measuring the degree of similarity between a pair of sentences; Spearman correlation between their cosine similarity and human judgments.
    \item \textbf{Summarization:} assessing machine-generated summaries by comparing them with human-written references; Spearman correlation between their cosine similarity and human ratings.
\end{itemize}

\section{Experimental settings}
\label{sec:settings}
\subsection{Model selection, embedding size, and quantization strategy}
Kusupati et al.~\cite{DBLP:conf/nips/KusupatiBRWSRHC22} demonstrated that models not trained using the MR loss do not perform well when a subspace of the embedding vector is used, and similar findings have been reported for 2D MRs~\cite{DBLP:journals/corr/abs-2402-14776}. Therefore, we selected four transformer-based text embedding models trained using either the 1D or 2D MR loss, with fewer than 1B parameters due to limited computational resources:
\begin{itemize}
    \item \textbf{Stella}: the \textit{stella\_en\_400M\_v5}\footnote{\url{https://huggingface.co/NovaSearch/stella_en_400M_v5}} model~\cite{DBLP:journals/corr/abs-2412-19048} is a transformer model with 400 million parameters, distilled from the \textit{gte-large-en-v1.5} and \textit{gte-Qwen2-1.5B-instruct} models and trained using the MR objective, with a default output dimensionality of 8192.
    \item \textbf{MBAI-1D}: the \textit{mxbai-embed-large-v1}\footnote{\url{https://huggingface.co/mixedbread-ai/mxbai-embed-large-v1}} model~\cite{DBLP:conf/acl/LiL24} is a transformer model with 335 million parameters, released by Mixedbread-AI. It is designed for English language tasks, has 24 layers, and a default embedding size of 1024. The model has been trained using the MR objective and contrastive learning on 700 million data pairs and over 30 million triplets using the AnglE objective. The company also notes that this model is well suited for binary embeddings~\cite{shakir2024binary}.
    \item \textbf{Nomic}: the \textit{nomic-embed-text-v1.5}\footnote{\url{https://huggingface.co/nomic-ai/nomic-embed-text-v1.5}} model~\cite{nussbaum2024nomic} is a transformer model with 137 million parameters, trained using the MR objective with a default output dimensionality of 768.
    \item \textbf{MBAI-2D}: the \textit{mxbai-embed-2d-large-v1}\footnote{\url{https://huggingface.co/mixedbread-ai/mxbai-embed-2d-large-v1}} model~\cite{DBLP:journals/corr/abs-2402-14776} is a transformer model with 335 million parameters, released by Mixedbread-AI. In contrast to the other models, this model also supports 2D MRs; besides the adjusted training objective, the model is similar to MBAI-1D in terms of architecture and training data.
\end{itemize}

For each model, we use eight hierarchical sizes, starting with the maximum output dimensionality of the model: $\mathbb{M} = \{8192, 4096, \dots, 64\}$ for Stella, $\{1024, 512, \dots, 8\}$ for MBAI-1D and MBAI-2D, and $\{768, 512, 256, \dots, 8\}$ for Nomic. For MBAI-2D, the only 2D MR model, we evaluate the intermediate layers 20, 16, and 12, based on the gradual degradation in performance observed in the original 2D MRs paper~\cite{DBLP:journals/corr/abs-2402-14776}, as well as the final layer 24.

Our experiments use (signed) int8 and binary quantization alongside the default float32 embeddings; in our implementation the conversion is performed by the \texttt{quantize\_\allowbreak embeddings} function of Sentence-Transformers. The conversion from floating-point to int8 is done using bucketing, which divides the range of each dimension, from its minimum to its maximum over the embeddings being quantized, into 256 equally sized buckets, with each bucket represented by a single int8 number; no separate calibration set is used. Conversion to binary numbers is done by setting the binary number to 1 if the floating-point number is positive and to 0 otherwise. The resulting bits are packed eight to a byte, so a binary embedding of size $m$ occupies $m/8$ bytes, in agreement with Eq.~\ref{eq:memory}. The packed codes are passed to the MTEB evaluators, which process them as ordinary numeric vectors; the Hamming-distance search used by dedicated binary indexes was not part of our protocol (Section~\ref{sec:limitations}).

\subsection{Dataset and task selection criteria}
\label{sec:tasks}
All tasks that are part of the MTEB subset used for the Hugging Face leaderboard were considered. Then, tasks that were too computationally expensive were excluded. Since seven model configurations (three 1D models and the 2D model with four layers) are evaluated for eight hierarchical sizes and three quantization techniques, resulting in a single task being repeated 168 times, we removed any task taking longer than 15~min on any model to complete in its default configuration. The resulting 38 tasks (Table~S1 of the Supplementary Material) span scientific literature, news, social media, technical forums, and customer service: classification (Banking77, Emotion, TweetSentimentExtraction, AmazonCounterfactual, MassiveIntent, MassiveScenario, MTOPDomain, MTOPIntent); clustering (the P2P and S2S variants of ArXivHierarchical, Biorxiv, Medrxiv, and StackExchange, together with Reddit and TwentyNewsgroups); pair classification (SprintDuplicateQuestions, TwitterSemEval2015, TwitterURLCorpus); re-ranking (AskUbuntuDupQuestions, MindSmallReranking, StackOverflowDupQuestions); retrieval (ArguAna, CQADupstackWebmastersRetrieval, NFCorpus); STS (BIOSSES, SICK-R, STS12 to STS17, STS22, STSBenchmark); and summarization (SummEval). Clustering groups either titles (S2S) or titles with abstracts or full posts (P2P) of scientific papers, forum posts, and news. Classification ranges from intent detection with up to 77 classes (Banking77) to emotion detection with six classes and counterfactual detection in product reviews. Pair classification detects duplicate questions and paraphrases; re-ranking orders candidate questions from technical forums and candidate news articles; retrieval covers counter-argument retrieval, duplicate questions from a webmasters forum, and medical documents. SummEval contains 1600 summaries of news articles paired with human expert ratings across four dimensions (coherence, consistency, fluency, relevance). The evaluations are performed on the respective test sets of the tasks, following the MTEB leaderboard methodology. Due to GPU memory problems, the Nomic model could not be evaluated on MedrxivClusteringP2P.v2, STS22, CQADupstackWebmastersRetrieval, and NFCorpus.

\paragraph{Impact of this filtering on the conclusions} This criterion is not neutral. Task runtime in MTEB is driven overwhelmingly by corpus size, so the large retrieval benchmarks, among them MSMARCO, Natural Questions, HotpotQA, DBPedia, FEVER, ClimateFEVER, QuoraRetrieval and TRECCOVID, are all excluded, and retrieval is represented by three of the smallest corpora. The global averages therefore under-weight the workload that embedding compression is most often deployed for. Moreover, as the number of candidate documents increases, the margin separating the correct document from its nearest distractors shrinks, so a given amount of compression noise is more likely to reverse a ranking. Our retrieval findings, already the most pessimistic of any category, should accordingly be read as a \emph{lower bound} on the degradation at larger scales.

\subsection{Software and hardware configuration}
All experiments are conducted using Python, with PyTorch, the Hugging Face Transformers and Sentence-Transformers libraries, and MTEB, in Jupyter notebooks, which can be accessed via a public GitHub repository\footnote{\url{https://github.com/marcopoli/Matryoshka-Quantization}}. The code is executed on Google Colab Pro, and all experiments are conducted on an \textit{NVIDIA Tesla T4 GPU with 16 GB of video memory}; all notebooks have 50 GB of system RAM available.

\section{Experimental results}
\label{sec:results}
We split the research question into the following sub-research questions for 1D and 2D MRs:
\begin{itemize}
    \item \textbf{RQ-1D-1 / RQ-2D-1:} How does the combination of 1D (2D) MRs and quantization affect the \textit{performance} across tasks?
    \item \textbf{RQ-1D-2 / RQ-2D-2:} How does the combination of 1D (2D) MRs and quantization affect the evaluation \textit{time} across tasks?
    \item \textbf{RQ-1D-3 / RQ-2D-3:} What is the trade-off between performance and \textit{per-vector storage} (and, for 2D MRs, evaluation time)?
\end{itemize}

\paragraph{A note on the timing results} Every evaluation time comes from a single MTEB run per configuration on a shared Google Colab runtime, whose run-to-run variability we could neither control nor measure. We therefore treat timing figures as indicative and interpret only timing effects that are large and consistent across configurations and task categories: sampling from an intermediate layer shortens evaluation substantially, and we find no evidence that int8 quantization brings a meaningful speed-up.

\subsection{Quantized 1D MRs}
\label{sec:results1d}
\begin{figure}[!htbp]
    \centering
    \begin{subfigure}{0.32\textwidth}\centering\includegraphics[width=\linewidth]{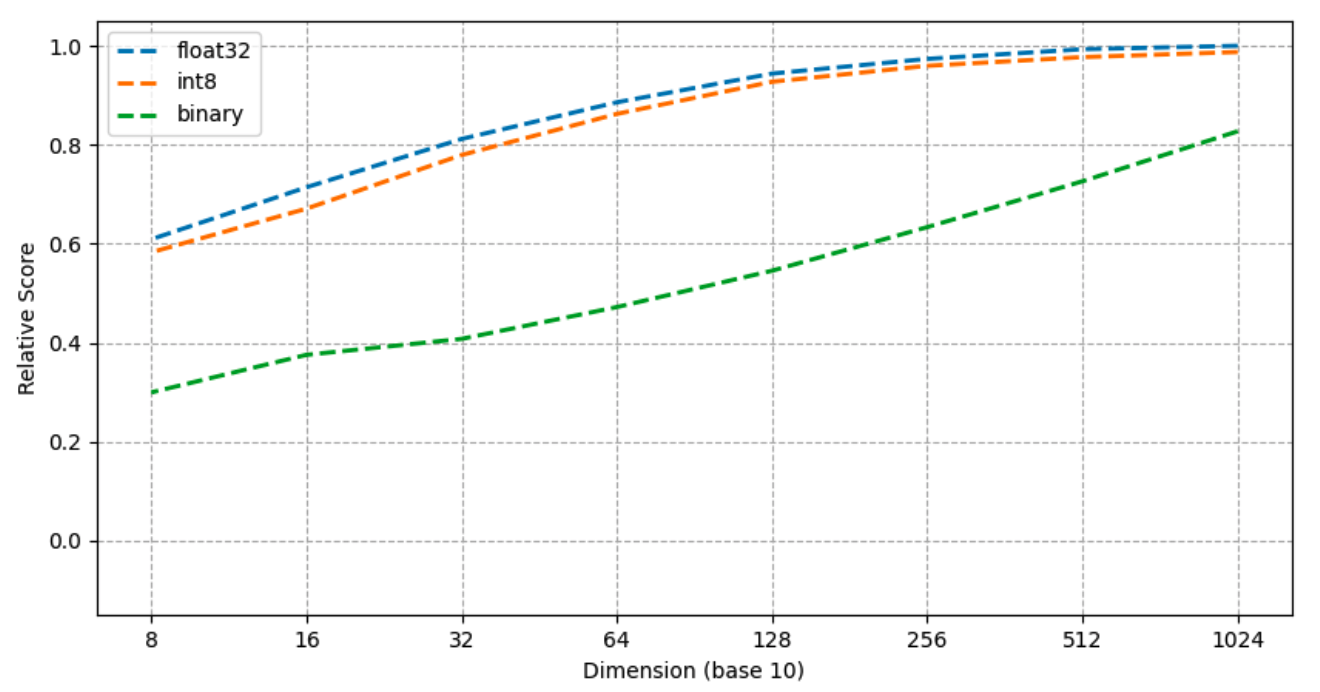}\caption{MBAI-1D: performance}\end{subfigure}\hfill
    \begin{subfigure}{0.32\textwidth}\centering\includegraphics[width=\linewidth]{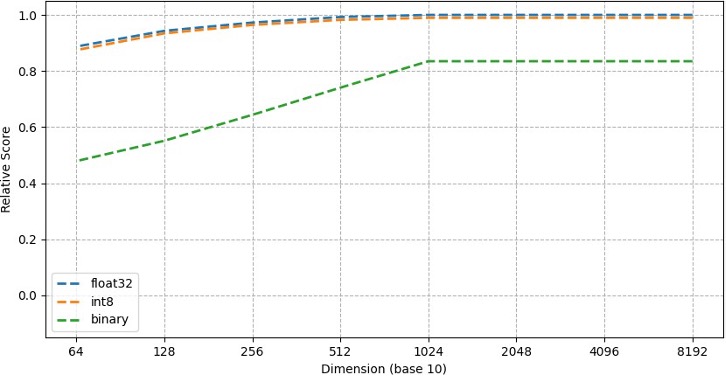}\caption{Stella: performance}\end{subfigure}\hfill
    \begin{subfigure}{0.32\textwidth}\centering\includegraphics[width=\linewidth]{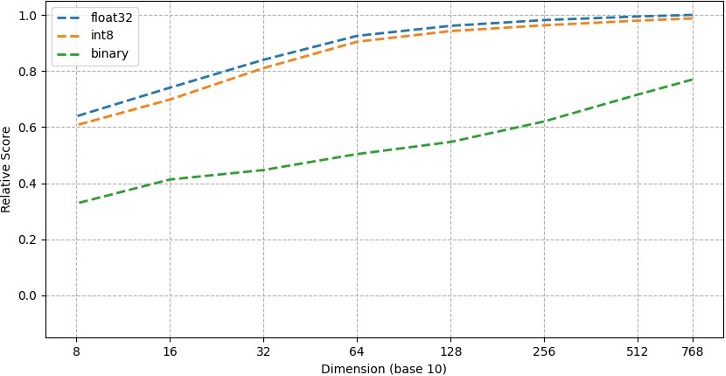}\caption{Nomic: performance}\end{subfigure}

    \vspace{4pt}
    \begin{subfigure}{0.32\textwidth}\centering\includegraphics[width=\linewidth]{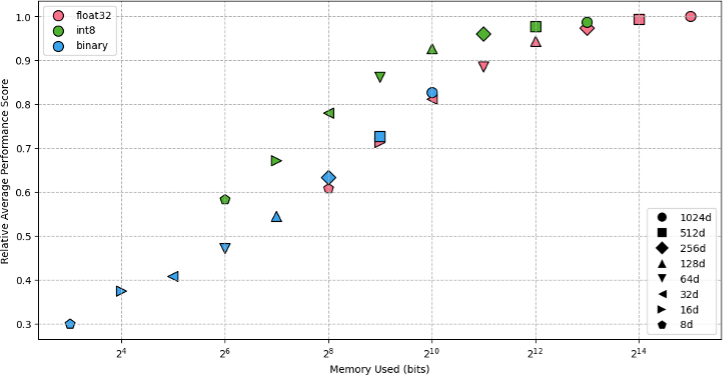}\caption{MBAI-1D: storage trade-off}\end{subfigure}\hfill
    \begin{subfigure}{0.32\textwidth}\centering\includegraphics[width=\linewidth]{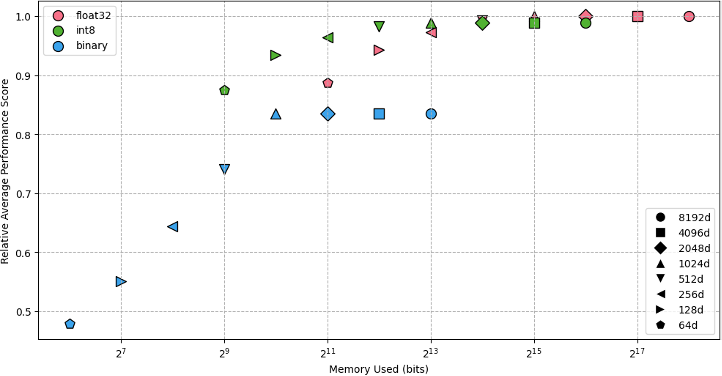}\caption{Stella: storage trade-off}\end{subfigure}\hfill
    \begin{subfigure}{0.32\textwidth}\centering\includegraphics[width=\linewidth]{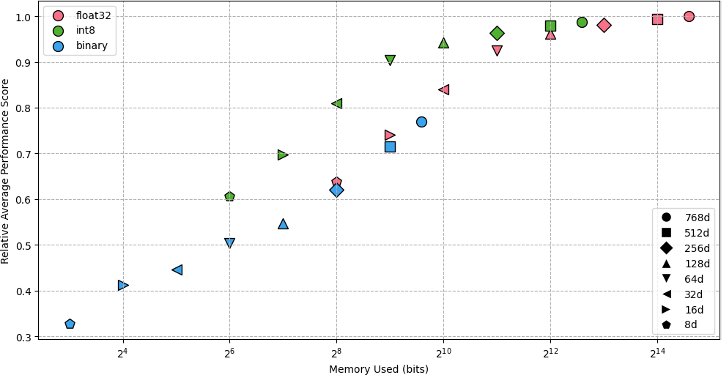}\caption{Nomic: storage trade-off}\end{subfigure}

    \vspace{4pt}
    \begin{subfigure}{0.32\textwidth}\centering\includegraphics[width=\linewidth]{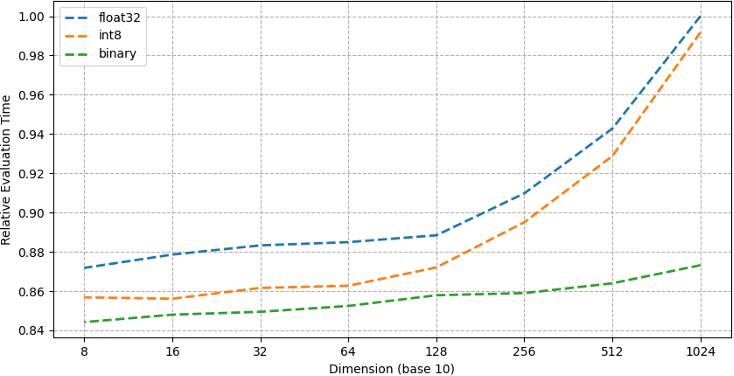}\caption{MBAI-1D: evaluation time}\end{subfigure}\hfill
    \begin{subfigure}{0.32\textwidth}\centering\includegraphics[width=\linewidth]{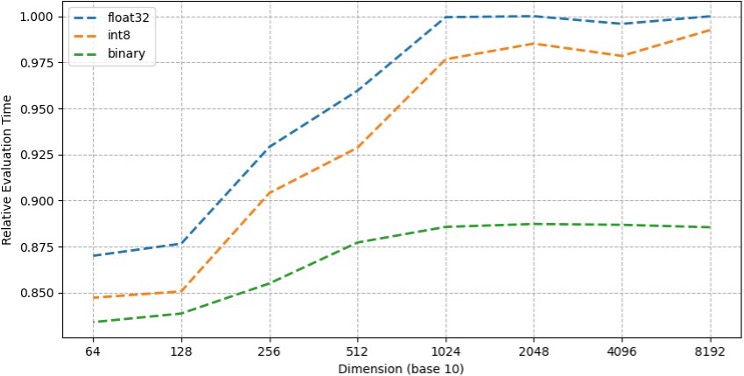}\caption{Stella: evaluation time}\end{subfigure}\hfill
    \begin{subfigure}{0.32\textwidth}\centering\includegraphics[width=\linewidth]{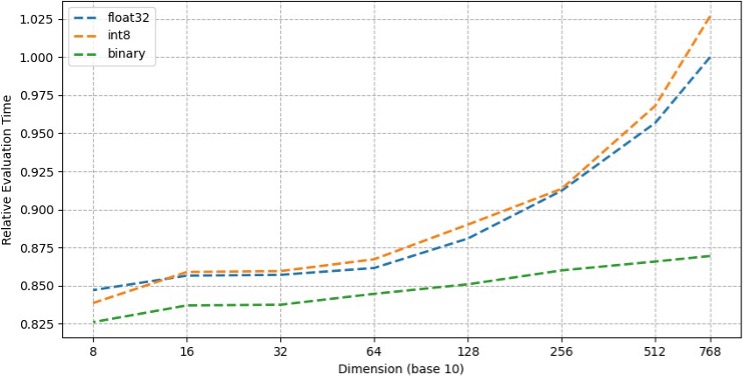}\caption{Nomic: evaluation time}\end{subfigure}
    \caption{Results for the 1D models over all selected MTEB tasks. (a)--(c) Average relative performance as a function of the hierarchical size for float32 (blue), int8 (orange), and binary (green) embeddings. (d)--(f) Average relative performance versus theoretical per-vector storage (bits, logarithmic scale); marker colors denote the precision and marker shapes the hierarchical size. (g)--(i) Average relative evaluation time as a function of the hierarchical size (single-run measurements). The reference point is each model's float32 baseline at full size.}
    \label{fig:results1d}
\end{figure}

\paragraph{Performance (RQ-1D-1)} Following the approach used in the MTEB leaderboard, we rely on the simple average of relative performance across all the selected tasks; this score is simple to understand, but the mean can give excessive weight to tasks with higher variance in their scores. Full results and tables are available in our GitHub repository\footnote{\url{https://github.com/marcopoli/Matryoshka-Quantization/blob/main/Results_Annexes.pdf}}. For the float32 embeddings of MBAI-1D (Fig.~\ref{fig:results1d}a), the reductions from 1024 down to 128 cost only 4.8\% of the original performance; as the MRs become smaller, the relative loss in performance increases, reaching 11\% at 64 dimensions. The int8 embeddings almost match the performance of the float32 ones, falling behind by only 6.6\% at most. The binary embeddings perform worse at all hierarchical sizes, and at $m = 1024$ the difference already exceeds 17\%.

Table~\ref{tab:categories}(a) reveals significant task-specific variations that are masked by the averaging process. Retrieval exhibits catastrophic performance degradation, dropping to just 0.06 at 8d for both float32 and int8 and completely failing (0.00) for binary embeddings, whereas Pair Classification maintains 70--71\% performance at 8d for float32/int8. Clustering shows excellent resilience with binary embeddings, maintaining 50\% performance at 8d compared to 57\% for float32 and int8. The int8 embeddings maintain performance very close to that of float32 across all tasks, with perfect parity at 1024d for Clustering, Pair Classification, and Retrieval. Nomic (Fig.~\ref{fig:results1d}c) supports all the observations made so far, while Stella (Fig.~\ref{fig:results1d}b) shows a slightly different behavior: the performance remains constant for all quantization techniques when reducing the output embedding size from 8192 to 1024, which suggests that the model effectively uses a dimensionality of 1024. Due to insufficient documentation of Stella's training procedure, we acknowledge this deviation without overemphasizing it.

\begin{table}[!htbp]
\centering
\small
\textbf{(a)} 1D MRs: MBAI-1D by hierarchical size and precision, relative to float32 at 1024 dimensions\\[2pt]
\resizebox{\ifdim\width>\linewidth\linewidth\else\width\fi}{!}{%
\begin{tabular}{@{}l cccc cccc cccc@{}}
\toprule
& \multicolumn{4}{c}{\textbf{Float32}} & \multicolumn{4}{c}{\textbf{Int8}} & \multicolumn{4}{c}{\textbf{Binary}} \\
\cmidrule(lr){2-5}\cmidrule(lr){6-9}\cmidrule(lr){10-13}
\textbf{Task category} & \textbf{1024d} & \textbf{256d} & \textbf{64d} & \textbf{8d} & \textbf{1024d} & \textbf{256d} & \textbf{64d} & \textbf{8d} & \textbf{1024d} & \textbf{256d} & \textbf{64d} & \textbf{8d} \\
\midrule
Average & \textbf{1.00} & 0.97 & 0.89 & 0.61 & 0.99 & 0.96 & 0.86 & 0.58 & 0.83 & 0.63 & 0.47 & 0.30 \\
Classification & \textbf{1.00} & 0.97 & 0.87 & 0.53 & 0.96 & 0.92 & 0.78 & 0.45 & 0.79 & 0.51 & 0.37 & 0.24 \\
Clustering & \textbf{1.00} & 0.97 & 0.85 & 0.57 & \textbf{1.00} & 0.97 & 0.84 & 0.57 & 0.79 & 0.59 & 0.52 & 0.50 \\
STS & \textbf{1.00} & \textbf{1.00} & 0.96 & 0.79 & 0.99 & 0.99 & 0.95 & 0.76 & 0.93 & 0.81 & 0.52 & 0.20 \\
Pair Class. & \textbf{1.00} & 0.99 & 0.97 & 0.70 & \textbf{1.00} & 0.99 & 0.97 & 0.71 & 0.95 & 0.83 & 0.53 & 0.33 \\
Retrieval & \textbf{1.00} & 0.92 & 0.67 & \underline{0.06} & \textbf{1.00} & 0.92 & 0.66 & \underline{0.06} & 0.49 & 0.13 & 0.04 & \underline{0.00} \\
Re-ranking & \textbf{1.00} & 0.98 & 0.91 & 0.74 & 0.99 & 0.97 & 0.92 & 0.72 & 0.88 & 0.76 & 0.65 & 0.29 \\
Summarization & \textbf{1.00} & 0.94 & 0.92 & 0.81 & 0.98 & 0.92 & 0.90 & 0.78 & 0.95 & 0.86 & 0.96 & 0.60 \\
\bottomrule
\end{tabular}}

\vspace{6pt}
\textbf{(b)} 2D MRs: MBAI-2D at layers 24 and 12 ($m = 1024$), relative to float32 at layer 24\\[2pt]
\resizebox{\ifdim\width>\linewidth\linewidth\else\width\fi}{!}{%
\begin{tabular}{@{}l ccc ccc ccc@{}}
\toprule
& \multicolumn{3}{c}{\textbf{Float32}} & \multicolumn{3}{c}{\textbf{Int8}} & \multicolumn{3}{c}{\textbf{Binary}} \\
\cmidrule(lr){2-4}\cmidrule(lr){5-7}\cmidrule(lr){8-10}
\textbf{Task category} & \textbf{L24} & \textbf{L12} & \textbf{$\Delta$} & \textbf{L24} & \textbf{L12} & \textbf{$\Delta$} & \textbf{L24} & \textbf{L12} & \textbf{$\Delta$} \\
\midrule
\textit{Average (all tasks)} & 1.000 & 0.862 & -0.138 & 0.986 & 0.853 & -0.133 & 0.849 & 0.738 & -0.111 \\
\midrule
Classification & 1.000 & 0.829 & -0.171 & 0.950 & 0.799 & -0.151 & 0.786 & 0.642 & -0.144 \\
Clustering & 1.000 & 0.951 & -0.049 & 0.994 & 0.945 & -0.049 & 0.831 & 0.839 & +0.008 \\
STS & 1.000 & 0.861 & -0.139 & 0.991 & 0.847 & -0.144 & 0.934 & 0.790 & -0.144 \\
Pair Class. & 1.000 & 0.872 & -0.128 & 1.000 & 0.874 & -0.126 & 0.949 & 0.799 & -0.150 \\
Retrieval & 1.000 & 0.602 & -0.398 & 1.002 & 0.619 & -0.383 & 0.632 & 0.291 & -0.341 \\
Re-ranking & 1.000 & 0.890 & -0.110 & 0.997 & 0.891 & -0.106 & 0.879 & 0.823 & -0.056 \\
Summarization & 1.000 & 0.901 & -0.099 & 1.008 & 0.933 & -0.075 & 0.946 & 0.897 & -0.049 \\
\bottomrule
\end{tabular}}

\vspace{6pt}
\textbf{(c)} 1D MRs: MBAI-1D relative evaluation time, relative to float32 at 1024 dimensions (single-run measurements)\\[2pt]
\resizebox{\ifdim\width>\linewidth\linewidth\else\width\fi}{!}{%
\begin{tabular}{@{}l cccc cccc cccc@{}}
\toprule
& \multicolumn{4}{c}{\textbf{Float32}} & \multicolumn{4}{c}{\textbf{Int8}} & \multicolumn{4}{c}{\textbf{Binary}} \\
\cmidrule(lr){2-5}\cmidrule(lr){6-9}\cmidrule(lr){10-13}
\textbf{Task category} & \textbf{1024d} & \textbf{256d} & \textbf{64d} & \textbf{8d} & \textbf{1024d} & \textbf{256d} & \textbf{64d} & \textbf{8d} & \textbf{1024d} & \textbf{256d} & \textbf{64d} & \textbf{8d} \\
\midrule
Average & 1.00 & 0.91 & 0.89 & 0.87 & 0.99 & 0.90 & 0.86 & 0.86 & 0.87 & 0.86 & 0.85 & \textbf{0.84} \\
Classification & 1.00 & 0.81 & 0.75 & \textbf{0.73} & 1.01 & 0.82 & 0.76 & 0.74 & 0.76 & 0.75 & \textbf{0.73} & \textbf{0.73} \\
Clustering & 1.00 & 0.83 & 0.77 & 0.75 & 1.02 & 0.83 & 0.77 & 0.76 & 0.80 & 0.76 & 0.75 & \textbf{0.73} \\
STS & 1.00 & 1.00 & \textbf{0.99} & 1.00 & 1.00 & 1.00 & 1.00 & 1.00 & 1.00 & \textbf{0.99} & 1.00 & \textbf{0.99} \\
Pair Class. & 1.00 & 0.96 & 0.95 & \textbf{0.94} & 1.03 & 0.97 & 0.95 & 0.95 & 0.96 & 0.95 & 0.95 & \textbf{0.94} \\
Retrieval & 1.00 & 0.99 & 1.00 & 1.00 & \textbf{0.89} & \textbf{0.89} & \textbf{0.89} & \textbf{0.89} & \textbf{0.89} & \textbf{0.89} & \textbf{0.89} & \textbf{0.89} \\
Re-ranking & 1.00 & 0.99 & 1.02 & 1.01 & 0.92 & 0.90 & 0.90 & 0.90 & 0.90 & 0.90 & 0.90 & \textbf{0.89} \\
Summarization & 1.00 & 1.00 & 1.05 & 1.04 & 0.92 & \textbf{0.91} & 0.91 & 0.91 & 0.91 & 0.91 & 0.92 & \textbf{0.90} \\
\bottomrule
\end{tabular}}
\caption{Relative performance and evaluation time by task category. (a) Bold marks the best value in each row; underlined values mark the collapse of retrieval at 8 dimensions. (b) The $\Delta$ columns give the change from layer 24 to layer 12 for each precision; the \textit{Average} row is computed over all individual tasks and therefore does not equal the unweighted mean of the category rows. (c) Bold marks the shortest time in each row; timings are indicative (Section~\ref{sec:results}).}
\label{tab:categories}
\end{table}

\paragraph{Evaluation time (RQ-1D-2)} With the caveat stated above, reducing the hierarchical size is associated with shorter evaluation times, most visibly for classification and clustering (up to 27\%; Table~\ref{tab:categories}c), for which downstream classifiers and clustering algorithms operate directly on the embeddings. Int8 embeddings are only up to 3 percent faster than float32 embeddings on average (Fig.~\ref{fig:results1d}g), and Nomic is the only model in which the int8 embeddings are not faster than the float32 ones for all hierarchical sizes (Fig.~\ref{fig:results1d}i), so we find no evidence of a meaningful speed-up from int8. Binary embeddings were faster than float32 ones in our runs, by a margin that we report as indicative. Retrieval runs with int8 and binary embeddings took 0.89 times the baseline time at every dimension, but the reduction is identical for every embedding size and for both precisions, while the float32 retrieval runs show no dependence on embedding size, which more plausibly reflects a single-run measurement than an effect of quantization. The absolute times are given in Figure~S1 of the Supplementary Material.

\paragraph{Storage trade-off (RQ-1D-3)} Fig.~\ref{fig:results1d}(d)--(f) plot per-vector storage against average relative performance. For all budgets between $2^{13}$ and $2^6$ bits, there are multiple combinations requiring the same amount of memory, for example float32 with $m = 256$ and int8 with $m = 1024$ at $2^{13}$ bits, and the int8 embeddings provide the best performance of all quantization techniques tested. For instance, the int8 representation for $m = 256$ achieves a relative performance of 96\% while only using $2^{11}$ bits, 16 times less than the baseline. The performance of binary embeddings is exceeded by int8 quantization for all memory budgets between $2^{10}$ and $2^6$ bits; with $m = 32$, 16, and 8 they reach budgets that no other configuration reaches, but they retain less than 50\% of the performance. These characteristics are consistent for all task categories except for classification, where float32 outperforms int8 at equal storage by almost 4.5\%; for all other tasks, int8 is at most 1.8\% worse than float32. For Stella, the flat behavior between 8192 and 1024 dimensions improves the trade-off: int8 with $m = 512$ only performs 1.8\% worse than float32 with $m = 8192$ while reducing the bits required by a factor of 64. Nomic shows the same characteristics as MBAI-1D.

\paragraph{Answers to 1D MRs} Concluding \textbf{RQ-1D-1}, both the reduction of embedding size and quantization lead to performance decreases whose magnitude varies by task category, and Retrieval is particularly vulnerable. Concluding \textbf{RQ-1D-2}, reducing the hierarchical size is associated with shorter evaluation times for classification and clustering, whereas we find no evidence that int8 is faster than float32. Answering \textbf{RQ-1D-3}, int8 embeddings present the best trade-off between performance and memory consumption, whereas binary representations offer an even greater reduction at the cost of a significant decrease in performance.

\subsection{Quantized 2D MRs}
\label{sec:results2d}
\begin{figure}[!htbp]
    \centering
    \begin{subfigure}{0.49\textwidth}\centering\includegraphics[width=\linewidth]{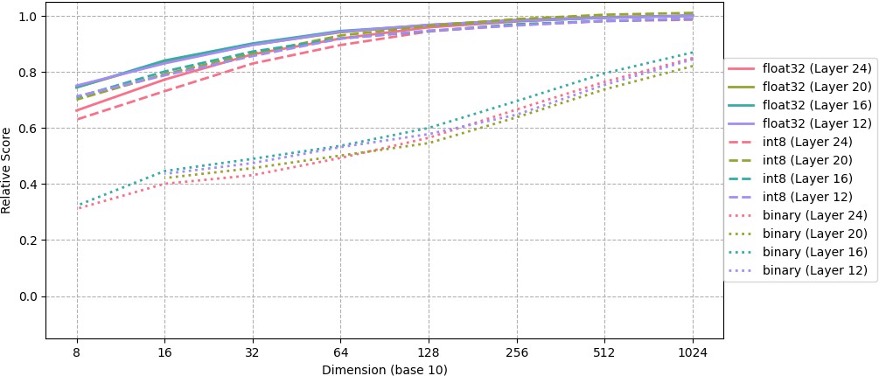}\caption{Average relative performance}\end{subfigure}\hfill
    \begin{subfigure}{0.49\textwidth}\centering\includegraphics[width=\linewidth]{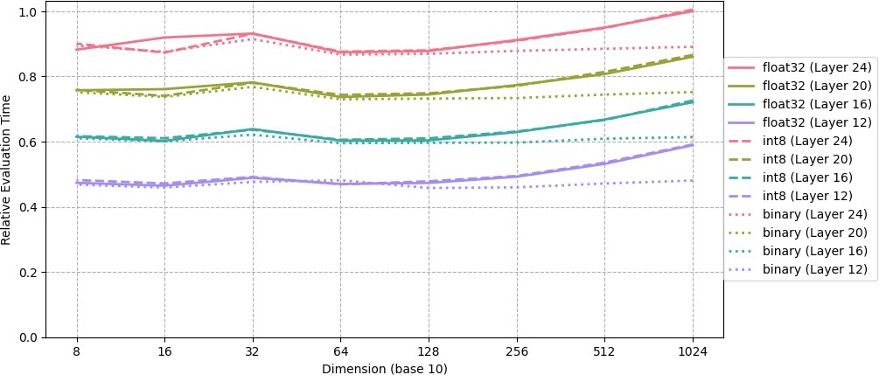}\caption{Average relative evaluation time}\end{subfigure}

    \vspace{4pt}
    \begin{subfigure}{0.56\textwidth}\centering\includegraphics[width=\linewidth]{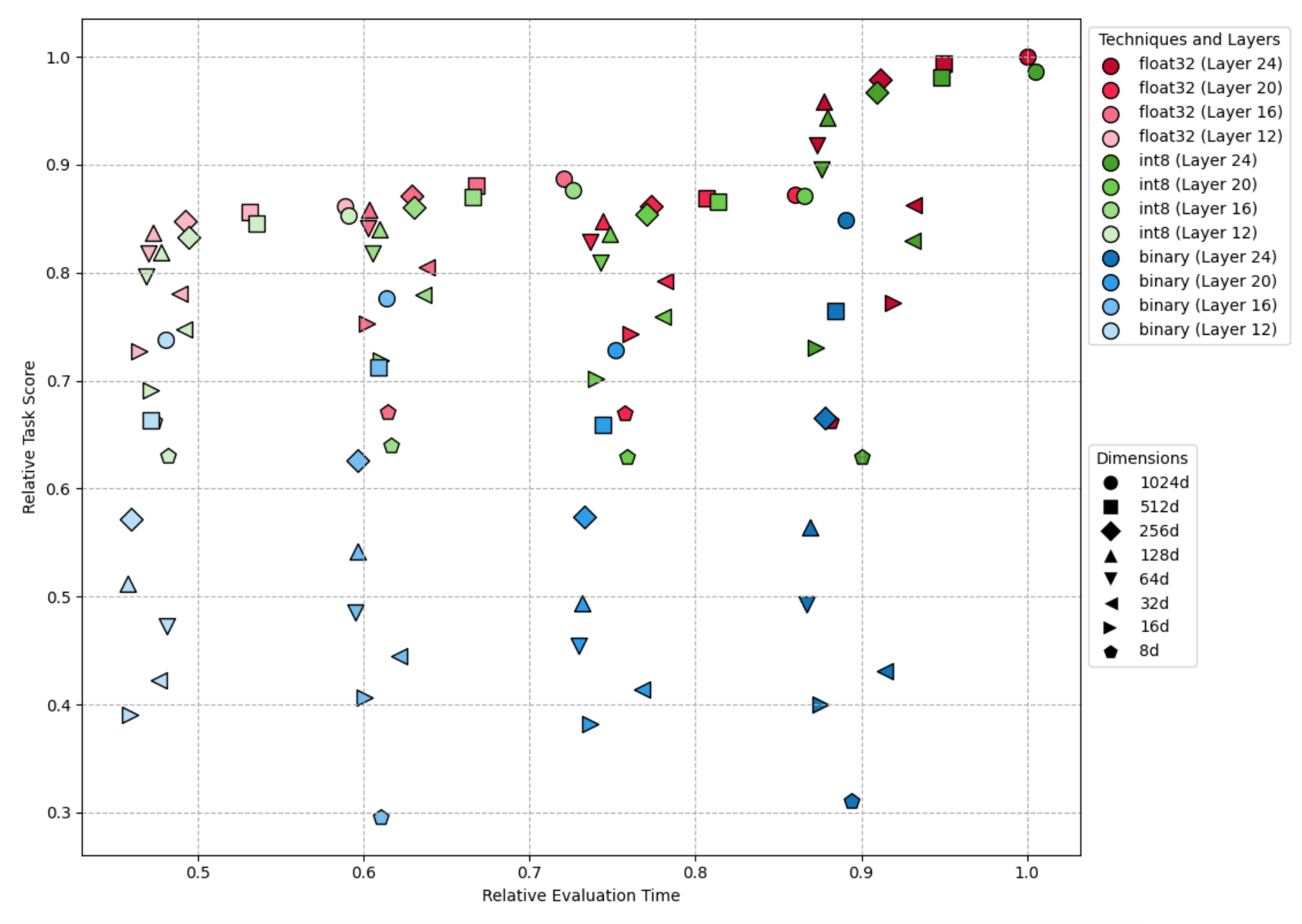}\caption{Trade-off among storage, evaluation time, and performance}\end{subfigure}
    \caption{Results for MBAI-2D over all selected MTEB tasks. In (a) and (b), line colors denote the inference layer (red: 24, yellow-green: 20, turquoise: 16, purple: 12) and line styles denote the precision (solid: float32, dashed: int8, dotted: binary). In (c), marker colors denote the precision and the inference layer, and marker shapes denote the hierarchical size; the reference point (1, 1) is the float32 baseline at full size from the final layer. Evaluation times are single-run measurements.}
    \label{fig:results2d}
\end{figure}

\paragraph{Performance (RQ-2D-1)} When only considering the final layer, the performance of the MBAI-2D model is within a 1\% margin of the MBAI-1D model for each combination of embedding size and quantization technique (Fig.~\ref{fig:results2d}a). Table~\ref{tab:categories}(b) compares layers 24 and 12 at $m = 1024$. Retrieval shows by far the largest degradation, losing 39.8\% for float32, 38.3\% for int8 and 34.1\% for binary, roughly three times the average loss of 13.8\%. At the opposite extreme, Clustering loses only 4.9\%, Summarization 9.9\%, Re-ranking 11.0\%, and Classification 17.1\%. This ordering is consistent with the 1D results, where Retrieval was likewise the category least tolerant of compression; we discuss the reason in Section~\ref{sec:why_retrieval}. The int8 embeddings track float32 closely across all categories, and the two cases in which they marginally exceed the float32 baseline at layer 24 (Retrieval at 1.002 and Summarization at 1.008) are within the noise. Categories already heavily damaged by binarization at layer 24, such as Retrieval at 0.632, have comparatively less left to lose, so their $\Delta$ values understate how unusable those configurations are. The degradation is also not strictly monotonic in layer depth: layer 16 ranks significantly better than layer 20 (Section~\ref{sec:stats}).

\paragraph{Evaluation time (RQ-2D-2)} The evaluation time decreases approximately linearly as the chosen layer for inference moves closer to the input layer (Fig.~\ref{fig:results2d}b), which is expected because the number of executed transformer blocks falls proportionally. The same decrease is observed for each task category, mirroring the findings of the 2D MRs paper~\cite{DBLP:journals/corr/abs-2402-14776}. The evaluation times for float32 and int8 embeddings are almost identical, binary embeddings take slightly less, and the hierarchical size has a much smaller effect than the layer.

\paragraph{Trade-off (RQ-2D-3)} In Fig.~\ref{fig:results2d}(c), between 0.87 and 1.0 on the x-axis only results from the 24th layer are found, between 0.7 and 0.87 only results from the 20th layer, and so on. Full-dimensional float32 embeddings inferred from layer 12 achieve a relative performance of 86.2\% while only requiring 59\% of the evaluation time. With int8 embeddings and $m = 128$, layer 12 maintains 81.9\% of the performance, requires only 47.8\% of the evaluation time, and reduces the memory consumption by a factor of 32; selected configurations are listed in Table~S2 of the Supplementary Material. Retrieval tasks provide the worst trade-off between performance and evaluation time, and clustering tasks the best.

\paragraph{Answers for 2D MRs} Concluding \textbf{RQ-2D-1}, the performance of the MBAI-2D model decreases as the layer used for inference moves closer to the input layer, with substantial task-specific variation, while the precision and size effects observed for 1D MRs hold at every layer. Concluding \textbf{RQ-2D-2}, the evaluation time decreases approximately linearly as the layer number decreases, and precision and size have much smaller effects. Concluding \textbf{RQ-2D-3}, given a moderate decrease in performance, the evaluation time was roughly halved in our measurements, while the per-vector storage was reduced by an exact factor of 32.

\subsection{Statistical significance analysis}
\label{sec:stats}
Averages alone cannot establish whether an observed difference is a real effect or an artifact of which tasks happen to be in the benchmark. Following Dem\v{s}ar~\cite{demsar2006statistical}, we use non-parametric tests, because relative performance is a bounded ratio whose distribution is neither normal nor homoscedastic across task categories, which also use different underlying metrics. For each of the three experimental factors, precision $q$, hierarchical size $m$, and inference layer $l$, we apply a Friedman test~\cite{friedman1937use} in which every combination of the remaining factors with the task category provides a matched block. Where the omnibus null hypothesis is rejected we apply the Nemenyi post-hoc test~\cite{nemenyi1963distribution}, with critical difference $\mathrm{CD} = q_{\alpha}\sqrt{k(k+1)/6N_b}$, and pairwise Wilcoxon signed-rank tests~\cite{wilcoxon1945individual} with Holm correction~\cite{holm1979simple}. All tests use $\alpha = 0.05$ and the full MBAI-2D grid of 96 configurations, the only model for which all three factors vary. Because the blocks share task categories they are not fully independent, so we rely on the effect sizes and the critical-difference verdicts, not on the magnitude of $p$. Table~\ref{tab:stats} summarizes the results; the pairwise Nemenyi comparison of the precisions is detailed in Table~S3 of the Supplementary Material.

\begin{table}[!htbp]
\centering
\footnotesize
\textbf{(a)} Omnibus Friedman tests and Nemenyi critical differences\\[2pt]
\begin{tabular}{@{}llcccccc@{}}
\toprule
\textbf{Factor} & \textbf{Blocks} & $k$ & $N_b$ & $\chi^2_F$ & $p$ & $W$ & $\mathrm{CD}$ \\
\midrule
Precision $q$ & category $\times$ $m$ $\times$ $l$ & 3 & 224 & 346.8 & $<10^{-15}$ & 0.774 & 0.221 \\
Hierarchical size $m$ & category $\times$ $q$ $\times$ $l$ & 8 & 84 & 511.9 & $<10^{-15}$ & 0.871 & 1.146 \\
Inference layer $l$ & category $\times$ $q$ $\times$ $m$ & 4 & 168 & 193.3 & $<10^{-15}$ & 0.383 & 0.362 \\
\bottomrule
\end{tabular}

\vspace{4pt}
\textbf{(b)} Per-category Wilcoxon signed-rank tests against float32 (Holm-corrected)\\[2pt]
\begin{tabular}{@{}lcccccc@{}}
\toprule
& \multicolumn{3}{c}{\textbf{int8 vs float32}} & \multicolumn{3}{c}{\textbf{binary vs float32}} \\
\cmidrule(lr){2-4}\cmidrule(lr){5-7}
\textbf{Task category} & $\overline{\Delta}$ & $p_{\text{Holm}}$ & Verdict & $\overline{\Delta}$ & $p_{\text{Holm}}$ & Verdict \\
\midrule
Classification & $-0.0751$ & $5.5\times10^{-6}$ & sig. & $-0.3734$ & $2.4\times10^{-6}$ & sig. \\
Clustering & $-0.0033$ & $1.8\times10^{-3}$ & sig. & $-0.2427$ & $2.4\times10^{-6}$ & sig. \\
STS & $-0.0212$ & $5.5\times10^{-6}$ & sig. & $-0.3352$ & $3.3\times10^{-9}$ & sig. \\
Pair class. & $-0.0001$ & $0.65$ & \textbf{n.s.} & $-0.3089$ & $3.3\times10^{-9}$ & sig. \\
Retrieval & $+0.0259$ & $6.7\times10^{-3}$ & sig. & $-0.3796$ & $3.3\times10^{-9}$ & sig. \\
Re-ranking & $-0.0036$ & $0.18$ & \textbf{n.s.} & $-0.2345$ & $3.3\times10^{-9}$ & sig. \\
Summarization & $-0.0303$ & $4.5\times10^{-4}$ & sig. & $-0.1142$ & $2.4\times10^{-6}$ & sig. \\
\bottomrule
\end{tabular}

\vspace{4pt}
\textbf{(c)} Hierarchical size: mean ranks and Nemenyi verdicts for one and two successive halvings ($\mathrm{CD} = 1.146$)\\[2pt]
\resizebox{\ifdim\width>\linewidth\linewidth\else\width\fi}{!}{%
\begin{tabular}{@{}l ccccccc@{}}
\toprule
Size $m$ & 1024 & 512 & 256 & 128 & 64 & 32 & 16 \\
Mean rank & 1.405 & 2.137 & 3.054 & 4.006 & 4.863 & 5.774 & 6.798 \\
\midrule
One halving & 512$\rightarrow$1024 & 256$\rightarrow$512 & 128$\rightarrow$256 & 64$\rightarrow$128 & 32$\rightarrow$64 & 16$\rightarrow$32 & 8$\rightarrow$16 \\
$|\Delta\bar{R}|$ (verdict) & 0.732 (n.s.) & 0.917 (n.s.) & 0.952 (n.s.) & 0.857 (n.s.) & 0.911 (n.s.) & 1.024 (n.s.) & 1.167 (sig.) \\
\midrule
Two halvings & 256$\rightarrow$1024 & 128$\rightarrow$512 & 64$\rightarrow$256 & 32$\rightarrow$128 & 16$\rightarrow$64 & 8$\rightarrow$32 & \\
$|\Delta\bar{R}|$ (verdict) & 1.649 (sig.) & 1.869 (sig.) & 1.810 (sig.) & 1.768 (sig.) & 1.935 (sig.) & 2.190 (sig.) & \\
\bottomrule
\end{tabular}}

\vspace{4pt}
\textbf{(d)} Inference layer: mean ranks ($\mathrm{CD} = 0.362$)\\[2pt]
\begin{tabular}{@{}lcccc@{}}
\toprule
Layer $l$ & 24 & 20 & 16 & 12 \\
Mean rank & 1.545 & 2.994 & 2.161 & 3.301 \\
\bottomrule
\end{tabular}

\vspace{4pt}
\textbf{(e)} Dispersion of relative performance across the seven task categories, with bootstrap 95\% confidence intervals of the category mean\\[2pt]
\resizebox{\ifdim\width>\linewidth\linewidth\else\width\fi}{!}{%
\begin{tabular}{@{}lccccc@{}}
\toprule
\textbf{Configuration} & \textbf{Mean} & \textbf{SD} & \textbf{Min} & \textbf{Max} & \textbf{95\% CI of the mean} \\
\midrule
float32, $m{=}1024$, L24 (baseline) & 1.000 & 0.000 & 1.000 & 1.000 & $[1.000,\ 1.000]$ \\
int8, $m{=}1024$, L24 & 0.992 & 0.019 & 0.950 & 1.008 & $[0.977,\ 1.002]$ \\
int8, $m{=}256$, L24 & 0.973 & 0.032 & 0.920 & 1.018 & $[0.951,\ 0.994]$ \\
int8, $m{=}128$, L24 & 0.950 & 0.047 & 0.881 & 1.002 & $[0.917,\ 0.981]$ \\
binary, $m{=}1024$, L24 & 0.851 & 0.115 & 0.632 & 0.949 & $[0.765,\ 0.921]$ \\
float32, $m{=}1024$, L12 & 0.844 & 0.113 & 0.602 & 0.951 & $[0.757,\ 0.905]$ \\
int8, $m{=}128$, L12 & 0.806 & 0.134 & 0.538 & 0.930 & $[0.706,\ 0.884]$ \\
binary, $m{=}512$, L12 & 0.653 & 0.232 & 0.172 & 0.846 & $[0.473,\ 0.781]$ \\
\bottomrule
\end{tabular}}
\caption{Statistical tests on the MBAI-2D grid of 96 configurations. (a) $k$ is the number of levels of the factor, $N_b$ the number of matched blocks, $W$ Kendall's coefficient of concordance, and $\mathrm{CD}$ the Nemenyi critical difference at $\alpha = 0.05$. (b) Tests over the 32 matched $(m, l)$ configurations of each category; $\overline{\Delta}$ is the mean difference in relative performance. (c), (d) Rank 1 denotes the best level; the mean rank at $m = 8$ is 7.964. (e) Intervals from 10{,}000 bootstrap resamples; they describe variability between task categories, not between repeated runs.}
\label{tab:stats}
\end{table}

\paragraph{All three factors matter, but not equally} All three omnibus nulls are rejected (Table~\ref{tab:stats}a), but Kendall's $W$ separates them: the orderings induced by precision ($W = 0.774$) and by hierarchical size ($W = 0.871$) are highly consistent from block to block, whereas the ordering induced by the inference layer is much less so ($W = 0.383$).

\paragraph{Precision: int8 is significant but small, binary is significant and large} All three pairwise differences exceed the critical difference ($\mathrm{CD} = 0.221$; mean ranks 1.306, 1.712, and 2.982 for float32, int8, and binary), but the float32--int8 gap corresponds to a mean difference in relative performance of just 1.5 percentage points, whereas the float32--binary gap corresponds to 28.4 percentage points. The confirmatory analysis on the MBAI-1D grid (7 categories $\times$ 4 dimensions, $N_b = 28$) is weaker still for int8: $\chi^2_F(2) = 43.75$, $p < 10^{-9}$, and the float32--int8 gap of 0.536 falls \emph{below} the critical difference of 0.626. Per category (Table~\ref{tab:stats}b), binary quantization is significantly worse than float32 in all seven categories, whereas int8 is \emph{statistically indistinguishable} from float32 for pair classification ($p_{\text{Holm}} = 0.65$) and re-ranking ($p_{\text{Holm}} = 0.18$), and its largest cost falls on classification ($-0.075$). For retrieval the mean difference is marginally \emph{positive} ($+0.026$), but the effect does not replicate on MBAI-1D ($-0.003$) and is best read as noise.

\paragraph{One halving of the dimension is free; two are not} No single halving produces a rank difference exceeding the critical difference (Table~\ref{tab:stats}c), with the sole exception of the final step from $m = 16$ to $m = 8$, whereas every two-step reduction is significant. Halving the embedding dimension once is therefore not a statistically detectable intervention, which makes compounding with int8 attractive, but reductions should be adopted one step at a time and re-validated.

\paragraph{Layer degradation is not monotonic} The mean ranks in Table~\ref{tab:stats}(d) order the layers as 24, 16, 20, 12; a monotonic account of depth would instead predict 24, 20, 16, 12. Layer 16 ranks significantly better than layer 20 ($|\Delta\bar{R}| = 0.833 > \mathrm{CD} = 0.362$), while layers 20 and 12 are statistically indistinguishable ($|\Delta\bar{R}| = 0.307$). The anomaly is concentrated in retrieval, where relative performance at layer 20 (0.513 at $m = 1024$, float32) falls well below that at layer 16 (0.718). The best intermediate layer must therefore be selected empirically per task.

\paragraph{Dispersion across task categories} The recommended int8 configurations are tightly clustered: int8 at $m = 256$ has a standard deviation of 0.032 across categories and a bootstrap 95\% confidence interval of $[0.951, 0.994]$ (Table~\ref{tab:stats}e). The aggressive configurations are the opposite: binary at $m = 512$ from layer 12 averages 0.653, but its category scores range from 0.172 to 0.846. This variability is \emph{between task categories}, not between repeated runs of the same configuration, because every configuration was executed once (Section~\ref{sec:limitations}).

\subsection{CO$_2$ emissions}
Since the exact locations of the servers used are unknown, we estimated the emissions based on the runtime, the maximum power consumption of the NVIDIA T4 GPU (70 watts), and the average grid carbon intensity of all Google data centers, 0.337~kg CO$_2$ equivalents per kWh\footnote{\url{https://cloud.google.com/sustainability/region-carbon}}. The experiments ran for 154.6 hours (27.3 for Stella, 24.6 for Nomic, 24.4 for MBAI-1D, and 78.3 for MBAI-2D), consumed about 10.8~kWh, and emitted 3.65~kg CO$_2$ equivalents in total (Table~S4 in the Supplementary Material). As expected, the tests for the MBAI-2D model emitted the most (1.85~kg), since all experiments were repeated for each layer considered for inference.

\FloatBarrier
\section{Discussion}
\label{sec:discussion}
The results answer the research question positively, with qualifications. Hierarchical representations and quantization can be combined: 1D MRs with int8 embeddings from the last layer ($m = 128$) reduce per-vector storage by an exact factor of 32 while retaining 94.4\% of the effectiveness. For those prioritizing \emph{speed}, however, only the 2D model offers a substantial lever, since we found no evidence that int8 is faster than float32. Moreover, the cost in effectiveness of any given setting varies strongly across task categories.

\subsection{Why some task categories are more sensitive than others}
\label{sec:why_retrieval}
A consistent finding across both the 1D and the 2D experiments is that retrieval degrades faster than any other category under both dimensionality reduction and quantization, while clustering and pair classification degrade the slowest. We identify four mechanisms, which act together.

\paragraph{1. Ranking is a harder task than assignment is} Classification, clustering and pair classification only require that an item fall on the correct side of a decision boundary, which is stable under perturbations that are small relative to the margin, whereas retrieval requires the correct ordering of candidates: any perturbation large enough to reverse two adjacent similarity scores changes the result. Re-ranking, which orders a short candidate list, sits between the two extremes.

\paragraph{2. Candidate-set size compresses the margins that must be preserved} As the number of documents grows, the expected similarity gap between the correct document and its nearest distractors shrinks, so a fixed amount of quantization noise becomes progressively more likely to reverse an ordering. A retrieval query is compared with thousands of documents, whereas an STS instance compares a single pair.

\paragraph{3. Some tasks can re-fit around the distortion; retrieval cannot} MTEB's classification tasks train a logistic-regression probe \emph{on the compressed embeddings themselves}, which can partially absorb a systematic distortion, and $k$-means re-estimates its centroids in the compressed space. Retrieval has no trainable component: the similarity function is fixed, so any distortion propagates directly into the final score. This explains why classification is robust to \emph{dimension} reduction, where the probe can compensate, but fragile under \emph{layer} reduction ($-17.1\%$ from L24 to L12), where the features the probe would need have not yet been computed.

\paragraph{4. Fine-grained discrimination lives in the trailing dimensions} Matryoshka training concentrates the dominant semantic directions in the leading coordinates, so the trailing coordinates carry the fine-grained distinctions needed to separate documents that are all topically relevant to a query. Truncation removes the tail first, and binary quantization also discards magnitude, which is why the two techniques interact so destructively for retrieval: at $m = 8$ with binary quantization, relative retrieval performance reaches 0.00.

Taken together, these mechanisms imply that the ordering is a property of the task structure rather than of the particular models or datasets used here. We therefore expect it to generalize, and we recommend treating the presence of a large-candidate-set ranking stage, rather than the choice of embedding model, as the primary signal that conservative compression settings are required.

\subsection{Practical guidelines for resource-constrained deployments}
\label{sec:guidelines}
Storage ratios are exact (Eq.~\ref{eq:memory}); effectiveness figures are relative to the $m = 1024$, float32, final-layer baseline, for which MBAI-1D and MBAI-2D differ by less than 1\%, and timings are indicative.
\begin{itemize}
    \item \textbf{High-effectiveness applications}, such as enterprise search: use $m = 1024$ or $m = 512$ with int8, which retain 99\% and 98\% of baseline effectiveness at 4$\times$ and 8$\times$ smaller per-vector storage. A single halving of the dimension is not a statistically detectable change, so moving from 1024 to 512 is close to free. The gain is capacity, not speed.
    \item \textbf{Balanced applications}, such as general-purpose RAG pipelines: deploy $m = 256$ with int8, which retains 96\% of baseline effectiveness at 16$\times$ smaller storage with a tight spread across task categories (SD 0.032). This is our recommended default when the workload is not known in advance.
    \item \textbf{Resource-constrained environments}, such as mobile applications: use $m = 128$ with int8, accepting 5.6\% degradation for a 32$\times$ storage reduction. This is safe for classification, clustering, STS and pair classification, but \emph{not} for first-stage retrieval, which retains only 0.89 of baseline at this setting on our small corpora and would degrade further at scale. Binary quantization costs 17\% of effectiveness even at full size and is significantly worse than float32 in every category; we recommend it only as the candidate-generation stage of a two-stage pipeline with Hamming-distance search and higher-precision rescoring~\cite{shakir2024binary}.
    \item \textbf{Speed-critical applications}: only the 2D model offers a substantial lever. Sampling from layer 12 combined with $m = 128$ and int8 roughly halved evaluation time in our measurements, at an 18.1\% cost in effectiveness, but the layer must be validated empirically rather than chosen by depth, since layer 16 ranked significantly better than layer 20, and the cost of layer reduction ranges from 4.9\% for clustering to 39.8\% for retrieval.
\end{itemize}
The appropriate lever also depends on the task category. Retrieval is highly sensitive to aggressive optimization, so moderate configurations ($m \geq 256$ with int8) are recommended; in RAG systems, however, the re-ranking component is markedly more resilient to layer reduction than first-stage retrieval (11.0\% against 39.8\% from layer 24 to layer 12 with float32), which makes 2D MRs a sensible optimization for the re-ranking stage. Clustering shows exceptional resilience to layer reduction, losing only 4.9\% when moving from layer 24 to layer 12, which makes it the strongest candidate for aggressive speed optimization; it also retains reasonable performance under binary quantization (50\% retention at 8 dimensions). Classification is highly tolerant of dimension reduction but less tolerant of layer reduction (17.1\%), so the recommended lever is truncation. STS and pair classification show excellent resilience to dimension reduction, and pair classification is also comparatively robust to layer reduction (12.8\%); since a single halving costs no measurable effectiveness, truncation is again the safer first step.

\paragraph{Hardware and systems considerations} First, \emph{the storage reduction is real and the speed reduction largely is not}: in our protocol the encoder forward pass dominates the cost, and the benchmark's evaluators do not exploit integer arithmetic, so int8 should be adopted for capacity, not latency. Second, the speed-ups reported for binary retrieval~\cite{shakir2024binary} come from Hamming distance over the packed words, a regime our protocol does not exercise, so a deployment that uses such search should expect a better speed return than our figures suggest. Third, \emph{intermediate-layer inference is the only lever that reduces embedding \emph{generation} cost}, since it alone changes how many transformer blocks are executed; deployments that continuously index new documents will find it more valuable than the storage levers.

\paragraph{Use case A: enterprise RAG over a five-million-chunk document store} The binding constraint is that the vector index should fit in the RAM of a single commodity server; the uncompressed baseline requires about 20.5~GB. Because the workload is retrieval-dominated, our results forbid aggressive truncation, and $m = 256$ with int8 gives about 1.28~GB, a 16$\times$ reduction, at 0.92 of baseline retrieval effectiveness and 0.96 averaged over all categories. If recall at the top of the ranking is critical, binary codes ($m = 1024$, 0.64~GB) can be used for candidate generation, with the int8 vectors retained for rescoring of the shortlist.

\paragraph{Use case B: on-device intent classification for a mobile assistant} The binding constraints are a fixed application memory budget and a strict latency target, and the workload is classification over a fixed label set, which tolerates dimension reduction well because its logistic probe re-fits around the compressed embeddings. With $m = 128$ and int8, each cached utterance embedding takes 128 bytes rather than 4~KB, at 0.94 of baseline effectiveness on average. Moving inference from layer 24 to layer 12 roughly halved evaluation time in our measurements, but classification loses 17.1\% at layer 12, so the layer should be validated empirically.

\subsection{Performance comparison with state-of-the-art MTEB models}
Among embedding models up to 1B parameters on the MTEB leaderboard (27 July 2025), the models studied here ranked 20th (Stella), 29th (MBAI-1D), 43rd (Nomic), and 191st (MBAI-2D) (Table~S5 in the Supplementary Material). The leading model, Qwen3-Embedding-0.6B, also supports Matryoshka Representations, as does the 8B-parameter Qwen3 model, which ranked first among embedding models under 10B parameters. Due to their later release, we were not able to conduct experiments on the newer models; however, we expect similar efficiency gains given their MR support.

\section{Limitations}
\label{sec:limitations}
\paragraph{Scope of the efficiency claims} The compression ratios reported in this paper describe theoretical per-vector storage, not resident system memory, and our timing figures are end-to-end MTEB task wall-clock times, not per-query latency.

\paragraph{Absence of controlled alternative-compression baselines} We do \emph{not} provide measurements for float16, PCA projection, Faiss scalar quantization, or product quantization under our own protocol, and Table~\ref{tab:comp} positions these families only qualitatively. We therefore claim only that Matryoshka truncation combined with int8 is effective, that it requires no fitted projection or trained codebook, and that it composes multiplicatively with precision reduction, not that it is \emph{superior} to PQ or PCA at an equal compression ratio. A comparison against PQ at matched bit budgets is the single most valuable extension of this work.

\paragraph{Single-run measurement and the absence of run-to-run variance} Every configuration was executed once, so the confidence intervals in Table~\ref{tab:stats}(e) quantify variability across task categories, not the stability of a repeated measurement. The effectiveness measurements are deterministic up to the seeded stochasticity of $k$-means and of the logistic-regression solver, but the timing measurements were collected on Google Colab, a shared-tenancy environment that we could neither control nor observe. We therefore do not report any timing percentage as a headline result: the timing figures in this paper, including the 12\% reduction measured for int8 embeddings at $m = 128$ and the reduction of roughly one half obtained by sampling from layer 12, are single measurements and should be read as indicative.

\paragraph{Benchmark composition and task selection} The runtime criterion removes all the large retrieval benchmarks, so our retrieval results are a lower bound on the degradation at larger scales (Section~\ref{sec:tasks}). MTEB itself is predominantly English and general-domain, so our findings may not transfer to multilingual deployments or to specialized domains.

\paragraph{Binary quantization: a conditional rather than general recommendation} Binary quantization is significantly worse than float32 in \emph{every} task category and only modestly faster in our setting, so we do not recommend it as a standalone representation. Our binary results should, moreover, be read as a lower bound on the technique: the packed codes were processed as ordinary numeric vectors rather than compared by Hamming distance, and at very small sizes the packing leaves the evaluators almost nothing to work with (at $m = 8$ a code is a single byte).

\paragraph{Model coverage, hardware dependence, and reproducibility} Only four models were evaluated, only one of which (MBAI-2D) supports 2D representations, so every conclusion about layer selection rests on a single model. Inference-layer selection adds a configuration dimension that must be validated empirically per task, which increases the tuning burden of deploying 2D representations, and evaluating the 2D model consumed more compute than the three 1D models combined. Results were obtained on NVIDIA T4 hardware, and quantization support varies considerably across accelerators, so the timing results in particular may not transfer to other processors.

\section{Conclusion}
\label{sec:conclusion}
This paper reports an empirical benchmark of the interaction between hierarchical (Matryoshka) text representations and embedding quantization, covering 168 configurations per task across 38 MTEB tasks and four transformer-based embedding models. Its answer to the research question is that the two techniques can be combined to improve efficiency while largely preserving performance, provided that storage and speed are considered separately and that the compression level is chosen per task category. Our central quantitative result is that 8-bit quantization combined with an eightfold dimensionality reduction ($m = 128$) retains 94.4\% of baseline effectiveness while reducing per-vector storage by an exact factor of 32, and $m = 256$ with int8 retains 96\% of effectiveness at a 16$\times$ storage reduction, the setting we would recommend as a default. For 2D representations, layer 12 combined with int8 at $m = 128$ roughly halved end-to-end evaluation time in our measurements, at a cost of 18.1\% in effectiveness.

Three findings qualify these headline numbers. First, \emph{storage gains and speed gains do not travel together}: we found no evidence that int8 shortens end-to-end evaluation time, and the only lever in our study that reliably reduces computation is sampling from intermediate layers. Second, \emph{statistical testing separates the differences that matter from those that merely exist}: int8 is statistically indistinguishable from float32 for pair classification and re-ranking and costs at most 1.5 percentage points overall, while binary quantization is significantly worse in all seven task categories at a cost of roughly 28 points; no single halving of the embedding dimension is statistically detectable above $m = 16$; and effectiveness does not degrade monotonically with layer depth. Third, \emph{the global average is the wrong object to optimize}: moving from layer 24 to layer 12 costs clustering 4.9\% and retrieval 39.8\%, and at $m = 8$ with binary quantization retrieval collapses to zero while clustering retains half its effectiveness. Retrieval-heavy systems should stay at $m \geq 256$ with int8; classification and clustering workloads can be compressed considerably further.

The natural continuation of this work is to rank hierarchical truncation against product quantization, PCA, and float16 storage at matched bit budgets on the identical 38-task protocol, together with repeated timing runs on dedicated hardware and an extension to full-scale retrieval corpora, to billion-parameter embedding models, and to adaptive schemes that select precision and dimensionality per query.

\section*{CRediT authorship contribution statement}

\textbf{Kai Ponel}: Writing – original draft, Validation, Software, Resources, Methodology, Investigation, Formal analysis, Conceptualization. \textbf{Marco Polignano}: Writing – review \& editing, Writing – original draft, Validation, Supervision, Methodology, Investigation, Formal analysis, Conceptualization. \textbf{Erasmo Purificato}: Writing – review \& editing, Supervision, Resources. \textbf{Aylin Julia Temizkan}: Writing – review \& editing, Supervision, Methodology. \textbf{Marco Levantesi}: Writing – original draft, Visualization, Resources. \textbf{Giovanni Semeraro}: Writing – review \& editing, Supervision. \textbf{Ernesto William De Luca}: Writing – review \& editing, Supervision, Methodology.

\section*{Declaration of Generative AI and AI-assisted technologies in the writing process}
During the preparation of this work, the authors used Generative AI and Grammarly AI in order to improve English language readability, correct grammatical errors, and simplify complex sentences. After using these tools/services, the authors reviewed and edited the content as needed and take full responsibility for the content of the publication.

\section*{Declaration of competing interest}
The authors declare the following financial interests/personal relationships which may be considered potential competing interests: Marco Polignano reports that financial support was provided by the Ministry of Education and Merit. Giovanni Semeraro reports that financial support was provided by the Ministry of Education and Merit. Kai Ponel reports a relationship with IBM Germany Research and Development that includes non-financial support. Aylin Julia Temizkan reports a relationship with IBM Germany Research and Development that includes employment. If there are other authors, they declare that they have no known competing financial interests or personal relationships that could have appeared to influence the work reported in this paper.

\section*{Acknowledgements}
This work is partially funded by Next Generation EU: NRRP Initiative, Mission 4, Component 2, Investment 1.3 -- Partnerships extended to universities, research centers, companies, and research D.D. MUR n. 341 del 15.03.2022 -- Next Generation EU (PE0000013 -- ``Future Artificial Intelligence Research -- FAIR'' - CUP: H97G22000210007).

\section*{Appendix A. Supplementary data}
Supplementary data to this article can be found online at doi:10.1016/j.ins.2026.124271.

\section*{Data availability}
All data used in this research are provided by the MTEB Python Library (\url{https://github.com/embeddings-benchmark/mteb}) which is maintained by third parties.

{\small
\setlength{\bibsep}{0pt plus 0.3ex}

}

\end{document}